\documentclass{article}
\usepackage[preprint]{format/corl_2021} 
\usepackage{times}

\usepackage{multicol}
\usepackage[table,xcdraw]{xcolor}
\newcommand{\etal}{\emph{et al.}}
\usepackage{hyperref}
\hypersetup{bookmarksopen,bookmarksnumbered,
pdfpagemode=UseOutlines,
colorlinks=true,
linkcolor=blue,
anchorcolor=blue,
citecolor=blue,
filecolor=blue,
menucolor=blue,
urlcolor=blue
}

\usepackage[skip=4pt,font=small,labelfont=bf]{caption} 
\usepackage[skip=0cm,list=true,labelfont=it]{subcaption}

\usepackage{textcomp}
\usepackage{color}
\usepackage{mathtools}
\usepackage{amsmath}
\usepackage{fixme}
\usepackage{epsfig,graphicx,amsmath,amsfonts}
\usepackage{xcolor,import}
\usepackage{transparent}
\usepackage{upgreek}
\usepackage{lscape} 
\usepackage{tabularx}
\usepackage{wrapfig}
\usepackage{graphicx}

\newcommand\numberthis{\addtocounter{equation}{1}\tag{\theequation}}
\usepackage[normalem]{ulem}
\usepackage{multirow}
\usepackage{import}
\usepackage{array, booktabs}
\usepackage[ruled,linesnumbered]{algorithm2e}

\definecolor{burntorange}{rgb}{0.8, 0.33, 0.0}

\newcommand{\xxnote}[3]{}
\ifx\hidenotes\undefined
  \renewcommand{\xxnote}[3]{\color{#2}{#1: #3}}
\fi

\DeclareMathOperator*{\argmin}{arg\,min}

\DeclareMathOperator*{\diag}{diag}

\newcommand{\argminprob}[1]{\underset{#1}{\argmin}}

\newcommand{\expect}[2]{\mathbb{E}_{#1}\left[#2\right]}

\newcommand{\bbm}{\begin{bmatrix}}
\newcommand{\ebm}{\end{bmatrix}}

\newcommand{\T}[3]{\prescript{#2}{}{#1}_{#3}}

\newcommand{\policy}{\pi}

\newcommand{\policyopt}{\pi^{*}}
\newcommand{\policysample}{\pi_{\theta}}

\newcommand{\termcost}{\hat{q}}
\newcommand{\meanstep}{\alpha_{\mu}}
\newcommand{\covstep}{\alpha_{\sigma}}

\newcommand{\storm}{\textsc{STORM}\xspace}
\newcommand{\mppi}[0]{\textsc{MPPI}\xspace}
\newcommand{\cem}[0]{\textsc{CEM}\xspace}
\newcommand{\rrtconnect}[0]{\textsc{RRTConnect}\xspace}
\newcommand{\rrtstar}[0]{\textsc{RRTStar}\xspace}

\newcommand{\mmc}[0]{\textsc{MMC}\xspace}
\newcommand{\moveit}{\textsc{MoveIt!}\xspace}

\newcommand{\getstate}[0]{\textsc{get\_state}}
\newcommand{\samplecontrols}[0]{\textsc{sample\_controls}}
\newcommand{\generaterollouts}[0]{\textsc{generate\_rollouts}}
\newcommand{\updatedistribution}[0]{\textsc{update\_distribution}}
\newcommand{\nextcommand}[0]{\textsc{next\_command}}
\newcommand{\executecommand}[0]{\textsc{execute\_command}}
\newcommand{\shift}[0]{\textsc{shift}}

\title{STORM: An Integrated Framework for Fast Joint-Space Model-Predictive Control for \\ Reactive Manipulation}
\author{Mohak Bhardwaj\textsuperscript{1,2}, Balakumar Sundaralingam\textsuperscript{1}, Arsalan Mousavian\textsuperscript{1}, Nathan Ratliff\textsuperscript{1}, \\[0.2cm] \textbf{Dieter Fox\textsuperscript{1,2}, Fabio Ramos\textsuperscript{1,3}, Byron Boots\textsuperscript{1,2}} \\[0.2cm]
\textsuperscript{1}NVIDIA \hspace*{20pt} \textsuperscript{2}University of Washington \hspace*{20pt} \textsuperscript{3} University of Sydney }

\begin{document}
\maketitle
\begin{abstract}
 Sampling-based model-predictive control~(MPC) is a promising tool for feedback control of robots with complex, non-smooth dynamics, and cost functions. However, the computationally demanding nature of sampling-based MPC algorithms has been a key bottleneck in their application to high-dimensional robotic manipulation problems in the real world. Previous methods have addressed this issue by running MPC in the task space while relying on a low-level operational space controller for joint control. However, by not using the joint space of the robot in the MPC formulation, existing methods cannot directly account for non-task space related constraints such as avoiding joint limits, singular configurations, and link collisions. In this paper, we develop a system for fast, joint space sampling-based MPC for manipulators that is efficiently parallelized using GPUs.
 Our approach can handle task and joint space constraints while taking less than 8ms~(125Hz) to compute the next control command. Further, our method can tightly integrate perception into the control problem by utilizing learned cost functions from raw sensor data. We validate our approach by deploying it on a Franka Panda robot for a variety of dynamic manipulation tasks. We study the effect of different cost formulations and MPC parameters on the synthesized behavior and provide key insights that pave the way for the application of sampling-based MPC for manipulators in a principled manner. We also provide highly optimized, open-source code to be used by the wider robot learning and control community. Videos of experiments can be found at: \href{https://sites.google.com/view/manipulation-mpc}{https://sites.google.com/view/manipulation-mpc}
\end{abstract}

\section{Introduction }
\label{sec:introduction}

Real-world robot manipulation can greatly benefit from real-time perception-driven feedback control~\cite{bohg2017interactive, kappler2018real}. Consider an industrial robot tasked with stacking boxes from one pallet to another or a robot bartender moving drinks placed on a tray~\cite{luo2017robust}. In both cases, the robot must ensure object stability via perception while respecting joint limits, avoiding singular configurations and collisions, and handling task constraints such as maintaining orientation during transfer. This leads to a complex control problem with competing objectives 
that is difficult to solve in real-time. 

Classic approaches to solving these tasks rely on operational-space control (OSC)~\cite{nakanishi2008operational,cheng2018rmpflow,dietrich2012reactive}, where the different task costs are formulated in their respective spaces and then projected into the joint space (i.e., the control space of the robot) via a Jacobian map. OSC methods are inherently local as they only optimize for the next time step while ignoring future actions or states.

MPC based approaches attempt to find a locally-optimal policy over a finite horizon starting from the current state using a potentially imperfect dynamics model. An action from the policy is executed on the system and the optimization is performed again from the resulting next state which can overcome the effects of model-bias. MPC has been successfully applied on real robotic systems allowing them to rapidly react to changes in the environment~\cite{bangura2014real, erez_humanoidmpc_2013, sciana_qphumanoidmpc_2020,cheng2018rmpflow, williams2016aggressive}. Existing MPC methods that operate in the joint space of a manipulator are limited to gradient-based approaches ~\cite{tassa2014control,williams2017model} which require the cost and dynamics to be differentiable. 
However, manipulation tasks often involve discontinuous phenomena such as contact and complex cost terms that are hard to differentiate analytically.
Sampling-based methods such as Model-Predictive Path Integral~(\mppi)~\citep{williams2017information} and Cross Entropy-Method~(\cem) offer a promising alternative. 
Here, control sequences are sampled using a simple policy followed by rolling out the dynamics model to compute a sample-based gradient of the objective function. 
 These algorithms make no restrictive assumptions about the cost, dynamics or policy class, are straightforward to parallelize and can be effectively applied on highly dynamic systems~\cite{williams2016aggressive, wagener19a}. These properties have also been a major factor in the increased adoption of sampling-based MPC 
by the Model-based RL community in recent years~\citep{chua2018deep}

\begin{figure}
    \centering     
    \includegraphics[trim=0 0.0cm 0 0, clip,width=0.98\textwidth]{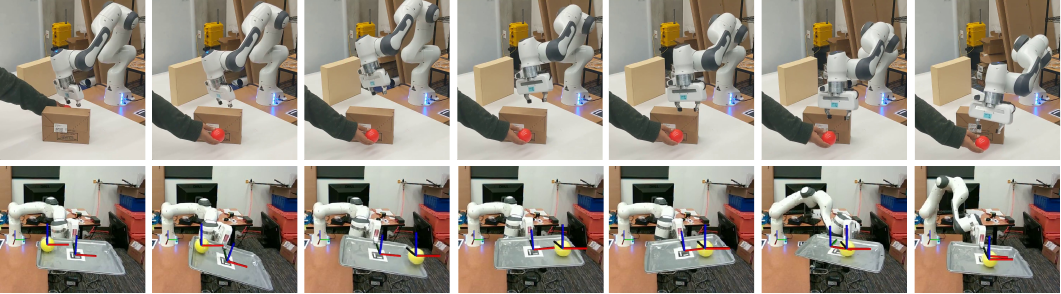}
    
  \caption{Our sampling-based model-predictive control framework operates in the joint space to enable a robot to achieve manipulation objectives such as tracking a moving target or balancing a ball on a plate while respecting constraints such as joint limits, singularity avoidance and collision avoidance via learned collision checking from raw sensor data. \vspace{-4mm}}
  \label{fig:intro}
  \end{figure}

However, a key question still remains to be answered - \textit{how well do these control algorithms transfer to high-dimensional robots like manipulators?} In this work, we describe an integrated system for sampling-based MPC that aims to answer this question. Our proposed framework, Stochastic Tensor Optimization for Robot Motion (\storm) implements a highly-parallelized control architecture that can optimize complex task objectives while simultaneously ensuring desirable properties such as smoothness, constraint satisfaction, and low control latency. 

A major criticism of sampling-based MPC algorithms for full joint space control has been their inability to produce smooth (low-jerk) trajectories~\cite{nagabandi2020deep}. To address this challenge, we study different sub-components of sampling-based MPC and make several novel contributions 
such as low discrepancy action sampling, smooth interpolation and cost-function design, and demonstrate their effectiveness in scaling sampling-based methods to real robot manipulators. 
We also demonstrate that \storm can incorporate learned components in the control loop, by using learned self and environment collision costs. 
We integrate our system on a real Franka Panda robot arm where we demonstrate that feedback driven sampling-based MPC is able to solve complex and dynamic manipulation tasks with simple models. In summary, our major contributions are
\begin{enumerate}


\item  A novel sampling-based MPC with the introduction of low discrepancy sampling, smooth trajectory generation and behavior-based cost functions that are key for producing smooth, reactive, and precise robot motions.

\item A feedback control framework that directly integrates learned perception components into the control loop in the form of a learned self collision cost and a discrete collision checker between the robot links and raw environment pointcloud from~\cite{danielczuk2020object}.

\item An open-source and highly optimized implementation of sampling-based joint-space MPC, which achieves a control rate of 125Hz on a single GPU, a speedup of 100x compared to existing MPPI based manipulation implementations~\cite{danielczuk2020object}.
\item Empirical evaluation in simulation and a real-world Franka Panda robot on dynamic control tasks.
\end{enumerate}


\section{Problem Definition}
\label{sec:problem_formulation}
We consider the problem of generating a feedback control law (or policy)  for a robot manipulator with $d$ joints performing user-specified tasks in an unstructured environment where it is subject to non-linear constraints and must react in real-time to overcome errors due to its internal dynamics model, state estimation, and perception. 
The problem can be modelled as optimal control of a discrete-time stochastic dynamical system described by the equation,~$x_{t+1} \sim P(x_{t+1} | x_t, u_t)$, where~$P(x_{t+1} | x_t, u_t)$ defines the probability of transitioning into state $x_{t+1}$ conditioned on $x_t$ and control input $u_t \in \mathbb{R}^d$. The robot chooses controls using a deterministic or stochastic closed-loop policy $u_t \sim \policy\left(\cdot | x_t \right)$, incurs an instantaneous cost $c(x_t, u_t)$ and transitions into the next state, and the process continues for a finite horizon $T$. The goal is to design a policy that optimizes a task-specific objective function,~
$
	\policyopt = \argminprob{\pi \in \Pi} \expect{\pi, P}{\sum_{t=0}^{T-1}  c(x_t, u_t)}
$
where $\Pi$ is the space of all policies. 

The above setup is akin to optimizing a finite horizon Markov Decision Process (MDP) with continuous state and action spaces as done in reinforcement learning approaches~\citep{6614995}. Solving for a complex globally optimal policy is a hard problem especially since the task objective $c$ could be sparse or difficult to optimize. MPC can be viewed as a practically motivated strategy that simplifies the overall problem by focusing only on the states that the robot encounters \textit{online} during execution and rapidly re-calculating a ``simple'' locally optimal policy. 
At state $x_t$, MPC performs a look-ahead for a shorter horizon $H<T$ using an approximate model $\hat{P}(\hat{x}_{t+1} | \hat{x}_t, a_t)$, approximate cost function $\hat{c}(x_t, a_t)$ and a parameterized policy $\pi_{\phi}$ to find the optimal parameters $\phi^{*}$ that minimize an objective function $L$

\begin{equation}
\label{eq:mpc_problem}
\phi^* = \argminprob{\phi} \; {L(\hat{c}, \termcost, \pi_{\phi}, \hat{P}, x_t)}
\end{equation}
where $\termcost(\cdot)$ is a terminal cost function that serves as a coarse approximation of the cost-to-go beyond the horizon. An action is sampled from $\pi_{\phi^{*}}$ and the optimization is performed again from the resulting state after applying the action to the robot. The optimization is hot-started from the solution at the previous timestep by using a \textit{shift} operator, which allows MPC to produce good solutions with few iterations of optimization (usually 1 in practice). 

In the next section, we first introduce a sampling-based MPC technique to solving the optimization in Eq.~\ref{eq:mpc_problem} and discuss the objective function, policy class, and update equations. We then present our approach for applying it to manipulation problems. An overview of the notation used in the paper is presented in Fig.~\ref{fig:notation_algorithm}.

\begin{figure}
\begin{minipage}[htp]{0.62\textwidth}
\scalebox{0.92}{
\centering
 \begin{tabularx}{\linewidth}{lX}
    \toprule
       \textbf{Variable}  & \textbf{Description} \\ \toprule
        $\theta_t,\dot{\theta}_t,\ddot{\theta}_t$ & joint position, velocity, acceleration\\
        $x_t=[\theta_t,\dot{\theta}_t,\ddot{\theta}_t]$ & robot state at time $t$ \\
        $u_t$ & joint space command at time~$t$\\ \midrule
        $h \in [0,H)$ & number of steps in horizon~$H$\\
        $n \in [0,N)$ & batch of control sequences\\
        $K$ & iterations of optimization\\
        $dt$ & change in time between  steps in horizon\\
        $\mathbf{dt}$ & vector of~$dt$ $[0,H)$ \\ \midrule
        $\mathbf{\phi}_t = \lbrace\phi_{t,h}\rbrace$ & parameters of policy at time t.  \\
        $\pi_{\phi_t}$ & distribution over control given state\\
        $\mathbf{\mu}_{t} = \mu_{t,h}$ & sequence of H Gaussian means \\
        $\mathbf{\Sigma}_t = \Sigma_{t,h}$  & sequence of H Gaussian Covariances \\
        $\mathbf{u} = \mathbf{u}_{n,h}$ & batch of N control sequences of length H \\
        $\mathbf{\hat{x}} = \mathbf{\hat{x}}_{n,h}$ & batch of N state sequences of length H \\
        $\mathbf{\hat{c}} = \mathbf{\hat{c}}_{n,h}$ & batch of N cost sequences of length H \\
        $L$ & MPC loss function \\ \bottomrule
        \end{tabularx}}
        \label{tab:notation}
\end{minipage}
\begin{minipage}[htp]{0.39\textwidth}
\centering
\scalebox{0.92}{
    \begin{algorithm}[H]
	 \caption{Sampling-Based MPC~\label{alg:mpc}}
	\SetKwInOut{Input}{Input}
	\SetKwInOut{Parameter}{Parameter}
	\SetKwInOut{Output}{Output}
	\Input{$\theta_{0},$}
	\Parameter{H, N, K}
	\For{$t=1 \ldots T$}{
	    $x_t \leftarrow$ \getstate()  \\ \label{line:get_state}
	    $\policysample \leftarrow$ \shift() \\ \label{line:shift_distribution}
	    \For{$i = 1 \ldots K$}{ 
		    $\mathbf{u} \leftarrow$ \samplecontrols($\pi_{\phi_{t}}, H, N$) \\ \label{line:sample_actions}
		    $\mathbf{\hat{x}}, \mathbf{\hat{c}}, \mathbf{\hat{q}} \leftarrow$ \generaterollouts($x_t, H$) \\ \label{line:rollouts}
		    $\phi_{t} \leftarrow$ \updatedistribution( $\mathbf{\hat{c}}, \mathbf{u}$) \\ \label{line:update distribution}
	    }
	    $u_t$ = \nextcommand($\pi_{\phi}$) \\ \label{line:get_next_action}
	    \executecommand($u_t$) \\ \label{line:execute_control} 
    }
\end{algorithm}
}
\end{minipage}
\caption{We summarize the notations used on the left and the sampling based MPC algorithm on the right.}
\label{fig:notation_algorithm}
\end{figure}

\section{Sampling-Based Model Predictive Control}
\label{sec:approach}
Sampling-based MPC iteratively optimizes simple policy representations such as time-independant Gaussians over open-loop controls with parameters $\phi_t$ such that $\pi_{\phi_t} = \prod_{h=0}^{H-1}\pi_{\phi_{t,h}}$. Here, $\mathbf{\phi}_t$ represent the sequence of means $\mathbf{\mu}_t=\left[\mu_{t,0} \ldots \mu_{t,H-1}\right]$ and covariances $\mathbf{\Sigma}_t=\left[\Sigma_{t,0} \ldots \Sigma_{t,H-1}\right]$ at every step along the horizon $H$. A standard algorithm is shown in~Fig.~\ref{fig:notation_algorithm}. At every iteration, the optimization proceeds by sampling a batch of $N$ control sequences of length $H$,  $\mathbf{u}_{n\in[0,N),h\in[0,H)}$, from the current distribution (Line~\ref{line:sample_actions}), followed by rolling out the approximate dynamics function using the sampled controls to get a batch of corresponding states $\mathbf{\hat{x}}_{n\in[0,N),h\in[0,H)}$ and costs $\mathbf{\hat{c}}_{n\in[0,N),h\in[0,H)}$ (Line~\ref{line:rollouts}). The policy parameters are then updated using a sample-based gradient of the objective function (Line~\ref{line:update distribution}).
After $K \geq 1$ optimization iterations we can either sample an action from the resulting distribution or execute the first action from the mean (Line ~\ref{line:get_next_action}). 
We next describe how the distribution is updated. Consider the  function~$\hat{C}(\cdot)$,
\begin{align*}
\hat{C}(x_t,u_t) &= \sum_{h=0}^{H-2} \gamma^{h} \hat{c}(\hat{x}_{t,h}, u_{t,h})  + \gamma^{H-1}\termcost(\hat{x}_{t,H-1}, a_{t,H-1}). \numberthis
\end{align*}
where $\gamma \in [0,1]$ is a discount factor that is used to favor immediate rewards. A widely used objective function is the exponentiated utility or the risk-seeking objective,
\begin{align*}
L &= \expect{\pi_{\theta}, \hat{P}}{\exp\left(\frac{-1}{\beta} \hat{C}(x_t,u_t)\right) \bigg\rvert \hat{x}_{0} = x_{t}} \numberthis
\end{align*}
where $\beta$ is a temperature parameter. For this choice of objective, the mean and covariance are updated using a sample-based gradient as, 
\begin{align*}
\mu_{t,h} &= (1 - \meanstep) \mu_{t-1,h} + \meanstep \frac{\sum_{i=1}^{N} w_i u_{t,h}}{\sum_{i=1}^{N} w_i} \numberthis \\ 
\Sigma_{t,h} &= (1 - \covstep) \Sigma_{t-1,h} 
+ \covstep \frac{\sum_{i=1}^{N} w_i (u_{t,h} - \mu_{t,h})(u_{t,h} - \mu_{t,h})^\top}{{\sum_{i=1}^{N} w_i}} \numberthis
\end{align*}
where $\meanstep$ and $\covstep$ are step-sizes that regularize the current solution to be close to the previous one and, 
\begin{equation}
w_i = \exp \frac{-1}{\beta}\left( \sum_{h=0}^{H-2} \gamma^{h} \hat{c}(\hat{x}_{h,i}, a_{h,i}) + \gamma^{H-1}\termcost(\hat{x}_{H-1,i}, a_{H-1,i}) \right). 
\end{equation}
The update equation for the mean is the same as the well-known Model-Predictive Path Integral Control (MPPI) algorithm~\citep{williams2017information}.  We refer the reader to~\citep{wagener2019online} for the connection and derivation of update equations.
While covariance adaptation has previously been explored in the context of Path Integral Policy Improvement~\cite{stulp2012path} to automatically adjust exploration, standard implementations of MPPI on real systems generally do not update the covariance~\citep{williams2017information, wagener2019online}. However, we observed that updating the covariance leads to better performance with a fewer number of particles, such as stable behavior upon convergence to the goal.
Once an action from the updated distribution is executed on the system, the mean and covariance are shifted forward with default values appended at the end to warmstart the optimization at the next timestep (Line~\ref{line:shift_distribution}).

The above formulation of MPC offers the flexibility to extract different behaviors from our algorithm by tuning different knobs such as the choice of approximate dynamics, running cost, terminal cost, the policy class and parameters such as the horizon length, number of particles, step sizes and discount factor $\gamma$. Next, we switch our focus to the domain of robot manipulation and build our approach for sampling-based MPC by systematically evaluating the effects of a subset of key design choices on the overall performance of the controller.



\subsection{Approximate Model}
\label{sec:model}
The MPC paradigm allows us to effectively leverage simple models that are both computationally cheap to query and easy to optimize, as re-optimizing the control policy at every timestep can help to overcome the effects of errors in the approximate model. 
We leverage this error correcting property of MPC and utilize the kinematic model of the manipulator as our approximate transition model. Let the robot state be defined in the joint space as~$x=[\theta,\dot{\theta}, \ddot{\theta}]\in \mathbb{R}^{3d}$ and the commanded action be the joint acceleration~$u \in \mathbb{R}^d$. At every optimization iteration, we compute the state of the robot across the horizon~($\mathbf{\hat{x}} = [\Theta_{n,h}, \dot{\Theta}_{n,h}, \ddot{\Theta}_{n,h}]$, $h\in[0,H)$, $n\in[0,N)$) by integrating the sampled control sequences~$\mathbf{u}_{n\in[0,N),h\in[0,H)}$ from the robot's initial state $x_\text{init} = [\theta_\text{init}, \dot{\theta}_\text{init},\ddot{\theta}_\text{init}]$~(i.e., current state of the real robot). This can be efficiently implemented as a tensor product followed by a sum:
\begin{align*}
    \label{eq:tensor_dynamics}
  \ddot{\Theta} &=  \mathbf{u} & 
  \dot{\Theta} &= \dot{\theta}_{\text{init}} + S_l(1) \diag (\mathbf{dt}) \ddot{\Theta} & 
  \Theta &= \theta_\text{init} + S_l(1) \diag (\mathbf{dt}) \dot{\Theta} \numberthis 
\end{align*}
where~$S_l(1)$ is a lower triangular matrix filled with 1, and~$\mathbf{dt}$ is a vector of delta timesteps across the horizon. We use variable timesteps, with a smaller dt for the earlier steps along the horizon for higher resolution cost near the robot's current state and larger ones for later steps to get a better approximation of cost-to-go, similar to~\citep{erez_humanoidmpc_2013}. We intentionally write~$\Theta_{n,h}$ as~$\Theta$ to highlight the fact that we compute the states across the batch and horizon with a tensor operation without the need to iteratively compute the states across the horizon. This significantly speeds up the computation and is key to achieving the 8ms control latency.

Given a batch of states $\mathbf{\hat{x}}$, the Cartesian poses~$\mathbf{X} \in \mathbf{SE}(3)$, velocities~$\dot{\mathbf{X}}$, and accelerations~$\mathbf{\ddot{X}}$ of the end-effector are obtained by using forward kinematics~$FK(\Theta)$, the kinematic Jacobian~$J(\Theta)$, and its derivative ~$\dot{J}(\Theta)$ as
\begin{align*}
  &X = FK(\Theta) 
  &\dot{X} = J(\Theta) \dot{\Theta} \qquad \qquad  
  &\ddot{X} = \dot{J}(\Theta) \dot{\Theta} + J(\Theta) \ddot{\Theta} \numberthis\label{eq:jac-d}
\end{align*}
We can compute the forward kinematics in batch as they depend only on the current state. This provides an easily parallelizable formulation of the robot model that is amenable to GPU acceleration.

\subsection{Cost Function}
\label{sec:costs}
The cost function $\hat{c}(s,a)$ encodes high-level robot behavior directly into the MPC optimization. This can be viewed as a form of cost-shaping that allows MPC to achieve sparse task objectives while also satisfying auxillary requirements such as avoiding joint limits, ensuring smooth motions and safety. 
%
We consider cost functions that are a weighted sum of simple cost terms, where each individual term encodes a desired robot behavior that can be easily implemented in a batched manner. 

\subsubsection{Reaching Goal Poses}
Given the desired and current Cartesian poses for the end-effector ~$\T{X}{w}{g}, \T{X}{w}{ee} \in \mathbb{SE}(3)$ respectively in the world frame $w$, we compute a task-space cost term that penalizes their distance 
\begin{align*}
  \hat{c}_\text{pose}(\T{X}{w}{ee},\T{X}{w}{g}) &= ||\alpha_1 (I - \T{R}{w}{g}^\top  \T{R}{w}{ee})||_2 
  +  || \alpha_2( \T{R}{w}{g}^\top \T{d}{w}{ee} - \T{R}{w}{g}^\top \T{d}{w}{g}) ||_2 \numberthis
\end{align*}
where~$\T{R}{w}{ee}, \T{R}{w}{g}$ are the rotation components and~$\T{d}{w}{ee}, \T{d}{w}{g}$  are translation components of the current and goal end-effector pose respectively. 
The weight vectors~$\alpha_1, \alpha_2 \in \mathbb{R}^3$ allow us to weigh errors in different directions and orientations with respect to each other and can be set to different values to obtain qualitatively different behavior such as partial pose reaching or enforcing partial pose constraints. For example, we can set a high weight in~$\alpha_1$ along a desired axes to maintain the goal orientation throughout the motion of the robot as we demonstrate in our experiments.

\subsubsection{Stopping for Contingencies}
The finite horizon makes MPC myopic to events that can occur further in the future. Thus, it is desirable to ensure that the robot can safely stop within the horizon in reaction to events that might be observed at timestep~$H-1$, especially in dynamic environments. We encode this behavior by computing a time varying velocity limit~$\dot{\theta}_{max}\in \mathbb{R}^{H}$ for every timestep in the horizon based on a user-specifed maximum acceleration~$\ddot{\theta}_{max}$ and the time until~$H-1$. This means the joint velocity of the robot must allow it to come to a stop at the end of the horizon by applying the max acceleration. Any state that exceeds this velocity is penalized by a cost which is expressed as
\begin{align*}
  &\dot{\theta}_{max} = S_u(1) \ddot{\theta}_{max}dt 
  &\hat{c}_\text{stop}(\dot{\theta}_t) = \begin{cases}
    ||\dot{\theta}_{max,t} - |\dot{\theta}|||_2  & \text{if } \dot{\theta}_{max,t} - |\dot{\theta}| > 0.0\\
    0,              & \text{otherwise}
  \end{cases} \numberthis                      
\end{align*}
where~$S_u(1)$ is an upper triangular matrix filled with 1. 

\subsubsection{Joint Limit Avoidance}
Given minimum and maximum limits on joint state~$\theta_{min}, \theta_{max}$ respectively, we penalize the joint state~$\theta_t$ only if it exceeds a safety threshold defined by a~$k_{jl}$ ratio of its full range.
\begin{align*}
  &\hat{\theta}_{min} = \theta_{min} + k_{jl}(\theta_{max} - \theta_{min}) 
  &\hat{\theta}_{max} = \theta_{max} - k_{jl}(\theta_{max} - \theta_{min}) \\ 
  \hat{c}_\text{joint}(\theta_t) &= \begin{cases}
    ||\theta_t-\hat{\theta}_{min}||_2 & \text{if }{\theta_t<\hat{\theta}_{min}}\\
    ||\hat{\theta}_{max} - \theta_{t}||_2 & \text{else if }{\theta_t > \hat{\theta}_{max}}\\
    0 & \text{otherwise}
    \end{cases} \numberthis
    \end{align*}
where~$\hat{\theta}_{min}$, $\hat{\theta}_{max}$ adds a safety threshold from the actual bounds of the robot. In our experiments, we chose $k_{jl}=0.1$. 

\subsubsection{Avoiding Cartesian Local Minima}
\label{sec:manip_cost}
The manipulability score describes the ability of the end-effector to achieve any arbitrary velocity from a given joint configuration. It measures the volume of the ellipsoid formed by the kinematic Jacobian which collapses to zero at singular configurations. Thus, to encourage the robot to optimize control policies that avoid future kinematic singularities, we employ a cost term that penalizes small manipulability scores~\cite{klein1987dexterity,vahrenkamp2012manipulability} 
\begin{align*}
  \hat{c}_\text{manip}(\theta_t) &= \begin{cases}
    1.0-\sqrt{J(\theta_t)J(\theta_t)^\top}, & \text{if} \sqrt{J(\theta_t)J(\theta_t)^\top} < k_m \\
    0.0, & \text{otherwise}
    \end{cases} \numberthis
\end{align*}

\subsubsection{Self Collision Avoidance}
Computing self-collision between the links of the robot for a large number of configurations can be computationally expensive~\cite{Rakita-RSS-18,danielczuk2020object}. Hence, similar to previous approaches~\cite{Rakita-RSS-18, danielczuk2020object}, we train a neural network that predicts the closest distance~\footnote{Distance is positive when two links are penetrating and negative when not colliding.} between the links of the robot given a joint configuration~($\theta$). One difference in our approach is the use of positional encoding~(i.e.,$[\sin(\theta),\cos(\theta)]$) as we found this to improve the accuracy of the distance prediction~\cite{mildenhall2020nerf}. Our network~(which we call jointNERF) has three layers with $[256,128,64]$ neurons and ReLU activations. We compute a cost term as shown below,
\begin{align*}
    \label{eq:coll_cost}
  \hat{c}_{\text{self-coll}}(\theta_t) &=  \max(0,\text{jointNERF}(\theta_t))  \numberthis                      
\end{align*}

\subsubsection{Environment Collision Avoidance}
Safe operation in cluttered environments requires a tight coupling between perception and control for collision avoidance. Gradient-based approaches~\citep{zucker2013chomp} generally rely on either known object shapes or pre-computed signed distance fields that provide gradient information for optimization. However, our sampling-based approach can handle discrete costs and as such we explore collision avoidance without using signed distances. Specifically, we use a learned collision checking function from Danielczuk~\etal~\cite{danielczuk2020object} that operates directly on raw pointcloud data and  classifies if an robot link pointcloud~$pc_l$ is in collision with the environment pointcloud~$pc_{env}$ given the robot link's pose~$\T{X}{}{l}$. 
\begin{align*}
    \label{eq:coll_cost}
  \hat{c}_\text{coll}(pc_l,pc_{env}, \T{X}{}{l}) &=  \begin{cases}
    1,  & \text{if } collision, \\
    0,              & \text{otherwise.}
  \end{cases} \numberthis                      
\end{align*}

Finally our running cost function is given by
\begin{align*}
    \hat{c}(x_t, u_t) &= \alpha_p\hat{c}_\text{pose} + \alpha_s\hat{c}_\text{stop} + \alpha_{j}\hat{c}_\text{joint} + \alpha_{m}\hat{c}_\text{manip} + \alpha_c( \hat{c}_{\text{self-coll}} + \hat{c}_\text{coll})\numberthis
\end{align*}


\subsection{Sampling Strategy for Control Sequences}
\label{subsec:sampling_strategy}
The method used for sampling controls from the Gaussian policy can have a great impact on the convergence of the optimization and can help embed different desirable behaviors such as ensuring smoothness. Pseudorandom sequences typically used in Monte Carlo integration exhibit an undesirable clustering of sampled points which results in empty regions. Whereas low-discrepancy sequences, where {\em low discrepancy} refers to the degree of deviation from perfect uniform sampling, alleviate this problem by defining deterministic samplers that correlate each point to avoid groupings~\cite{niederreiter1992random}. Halton sequences~\cite{Halton64} are a widely used form of {\em low discrepancy} number generators that attempt to improve the rate of convergence of Monte Carlo algorithms, and are reported to achieve superior performance in high-dimensional problems~\cite{dick2013qmc}. In particular, the Halton sequence uses the first $p_i, \ldots, p_d$ prime numbers to define a sequence, $\mathbf{w}_1, \mathbf{w}_2, \ldots$, for integers $i\geq0$ and $b\geq2$ where $\mathbf{w}_i=(\phi_{p_1}(i), \ldots, \phi_{p_d}(i))$ and $\phi_b(i)=\sum_{a=1}^\infty i_a b^{-a}$, with $i=\sum_{a=1}^\infty i_a b^{a-1}$ for $i_0, i_1, \dots \in \{0,1,\ldots,b-1\}$
~\footnote{See \url{http://extremelearning.com.au/unreasonable-effectiveness-of}\\ \url{-quasirandom-sequences/} for a visualization of different low-discrepancy methods.}. We incorporate Halton sequences for sampling controls that can provide a better estimate of the objective function gradient. Controls from the Halton sequence are sampled once at the beginning and then transformed using the mean ($\mathbf{\mu}_t$) and Covariance $\mathbf{\Sigma}_t$ of the current Gaussian policy.

\begin{wrapfigure}[28]{r}{0.4\textwidth}
\vspace{-20pt}
\centering
    \includegraphics[width=\linewidth]{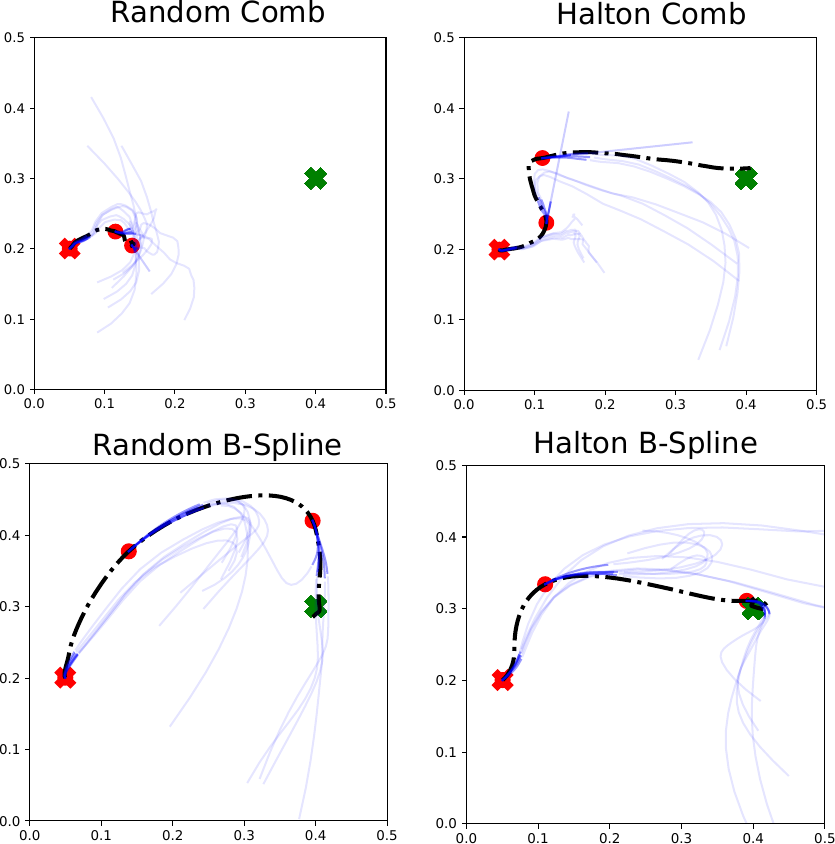}
\caption{We compare our sampling scheme on a planar robot reacher task~\cite{kobilarov2012cross}. The holonomic robot must move from its initial position~(green cross) to the desired position~(red cross) while avoiding obstacles shown as the grey regions. The path taken by the robot is given by the black dot-dashed line and red circles are positions of the robot after every 30 timesteps. The blue lines are the rolled out trajectories of the top 5 particles. Our Halton B-Spline is able to find a smooth short path to the goal while Random B-Spline takes a longer path. Halton with a comb filter is not smooth as shown by the sudden path changes.
    }
    \label{fig:halton_sample}
\end{wrapfigure}
Furthermore, we explore two different strategies for enforcing smoothness in sampled control sequences. The first method is a \textit{comb} filter that uses user-specified coefficients $\left[c_1, c_2, c_3\right]$ to filter out each sampled control trajectory along the horizon as $u_{t,h} = c_1 u_{t,h} + c_2 u_{t,h-1} + c_3 u_{t,h-1}$
This method has previously been used with sampling-based control techniques~\citep{yang2020data,summers2020lyceum}, however, it requires extensive tuning and the filtered trajectories are not guaranteed to be smooth as neighboring samples in the horizon can have large difference in magnitude.

We propose an alternate strategy to enforce smoothness by fitting B-splines of degree $3$ to controls sampled using a Halton Sequence. The resulting curve is sub-sampled at a finer resolution to obtain smooth joint acceleration samples which are then integrated to obtain corresponding joint velocity and position trajectories using Eq~\ref{eq:tensor_dynamics}. In Fig.~\ref{fig:halton_sample} we show a qualitative comparison between different combinations of sampling and smoothing strategies for a planar robot trying to reach a goal while avoiding obstacles where we see that our Halton + B-Spline sampling strategy is able to better explore the action space while maintaining smoothness (see supplementary for details). 

\textbf{Covariance Parameterization}: Conventionally, sampling-based MPC algorithms such as MPPI parameterize the covariance of the Gaussian policy to be of the form $\Sigma_{t,h} = \sigma_u * I_{d \times d}$ where $\sigma_u$ is a scalar value and $I_{d \times d}$ is a $d \times d$ identity matrix, which forces the covariance to be the same across all control dimensions. However, in the case of manipulators it is desirable to allow the covariance of different joints to adapt independently so they can potentially optimize different cost terms such as pose reaching versus increasing manipulability. Thus we also consider covariance of the form $\Sigma_{t,h}^{U} = \sigma_u^T I_{d \times d}$ where $\sigma_u = [\sigma_1, \ldots, \sigma_d]$. Each term in $\sigma_u$ is then adapted based on the rollouts.  Adapting the covariance along action dimensions has also been employed by~\citep{chua2018deep} for CEM. 

Our sampling strategy also offers us the flexibility of incorporating certain fixed set of action trajectories which could be task-dependent or even a library of pre-defined desired motions. We leverage this fact by incorporating a set of zero acceleration or ``null'' particles which allows the robot to coast at a constant velocity once the robot is accelerated sufficiently and also easily stop at the goal as demonstrated in our experiments.



\section{Experimental Evaluation}
\label{sec:results}

\begin{figure}
  \centering
  \includegraphics[width=0.98\linewidth,trim=0 2.99cm 0 0, clip]{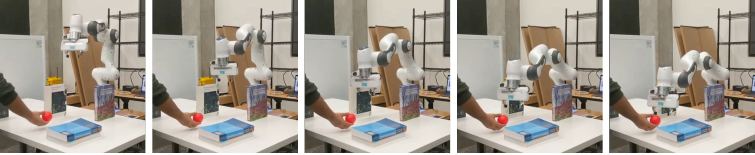}
  \caption{We show a sequence from our collision avoidance experiment where the robot has to move between thin walls to reach the orange ball held by the human. The robot tries to reach the ball but moves only as close as possible as any further motion would cause the elbow to hit the right book.} 
  \label{fig:coll_avoidance_real}
  \vspace{-4mm}
\end{figure}

Through our experiments we aim to analyze the effectiveness of \storm as a framework for real-time, perception-driven feedback control in real-world manipulation scenarios. To this end, we first study the performance of \storm in reacting to changing end-effector targets from perception data while satisfying task constraints such as maintaining user-specified orientation and avoiding obstacles in cluttered scenes. Second, we consider the dynamic task of balancing a ball on a tray grasped by the end effector that uses an approximate model of the ball dynamics. We provide results of ablation studies in simulation for different components of our framework such as cost terms, sampling strategy and policy parameterization in Appendix~\ref{sec:appendix:ablation_studies}, and a comparison to \moveit and OSC~\cite{haviland2020purely} for the standard pose reaching problem in Appendix~\ref{sec:appendix:reach_pose_exp}. Detailed results of ablations as well as a qualitative comparison to Reimannian Motion Policies (RMPs)~\citep{ratliff2018riemannian} can be found on our website~(\href{https://sites.google.com/view/manipulation-mpc}{https://sites.google.com/view/manipulation-mpc}).

\subsection{Tracking Moving Targets while Handling Task Constraints}
A key strength of feedback-based MPC over ``plan and execute'' and OSC approaches is its ability to simultaneously optimize complex cost functions over a long horizon while demonstrating reactive behavior. We demonstrate this by having the robot react to changing goal poses obtained from noisy perception, while satisfying task constraints such as maintaining desired orientation and avoiding obstacles. In these experiments, a ball held by a human is tracked using a depth camera and the robot tries to reach the ball as the human moves it to different locations in the workspace. 

\textbf{Obstacle Avoidance}: Demonstrating reactive motion and reasoning about obstacle avoidance in cluttered environments, while simultaneously coordinating large degrees of freedom leads to a hard online optimization problem. Standard planning and OSC approaches often fall short in optimizing for such behavior.

\begin{wrapfigure}[22]{r}{0.4\textwidth}
\centering
\vspace{-0pt}
    \includegraphics[width=0.4\textwidth, trim=0 0 0 0.1cm, clip]{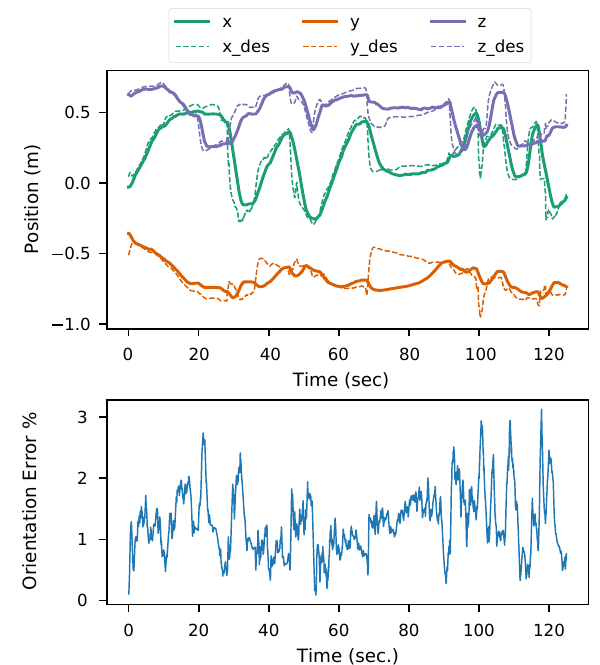}
    \caption{We move the ball across the workspace while having a high weight on maintaining a specific orientation of the end-effector. Our control scheme can maintain the orientation during motion as seen by the very low orientation error~($<3\%$).}
  \label{fig:orientation_constraint_real_error_plots}
\end{wrapfigure}

To test our method's capability to handle such scenarios, we setup two different table top environments, one consisting of two blocks representing common pick and place environments as shown in~Fig.~1 and an environment with thin walls to represent a densely occupied space, as shown in Fig.~\ref{fig:coll_avoidance_real}. We use our perception based ball tracker to make the robot reach different positions in the environment. 
During the experiment, we also move the ball to some positions that are not reachable by the robot due to possibility of collision between the robot's links and the obstacles. As seen in Fig.~\ref{fig:coll_avoidance_real}, the robot handles these situations very well as it prioritizes collision avoidance over reaching the pose accurately. We present our full obstacle avoidance experiments as well as several experiments in simulation in the accompanying videos.


\textbf{Orientation Constraints}: Several common manipulation tasks such as moving a filled cup require maintaining orientation during motion which reduces the feasibility region of a controller. We test this scenario by imposing orientation constraints on the end-effector while tracking the ball. Our controller achieves a median quaternion error of $1.2485 \%$ while tracking the ball with sufficient accuracy as shown in Fig.~\ref{fig:orientation_constraint_real_error_plots}. 
\vspace{-0.2cm}
\subsection{Dynamic Object Balancing}
\begin{wrapfigure}{r}{0.4\textwidth}
  \centering
    \includegraphics[width=0.9\linewidth]{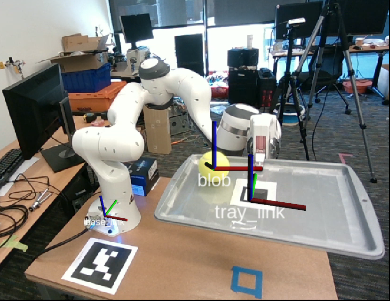}
    \caption{Snapshot of Ball Balancing Task}
    \label{fig:balance_snap}
    \vspace{-2mm}
\end{wrapfigure}

\vspace{-0.2cm}
We consider a hard dynamic manipulation task where the robot tries to balance a ball placed on a tray grasped by a parallel jaw gripper as shown in Fig.~\ref{fig:intro}. The location of the ball is tracked using the RGBD input from a RealSense camera at 30Hz. We use a simplified dynamics model of the object rolling on the tray under acceleration due to gravity and do not explicitly account for friction or inertial properties of the ball and do not perform any system identification.  Fig.~\ref{fig:balance_snap} shows a snapshot of the robot performing the task. The purpose of this task is to demonstrate the robustness of MPC under severe model-bias while optimizing complex objectives and dealing with noisy perception. We refer the reader to Appendix~\ref{sec:appendix:ball_balancing} for more details of the experimental setup and ball dynamics. 

\begin{figure}
    \centering
    \includegraphics[width=0.9\textwidth]{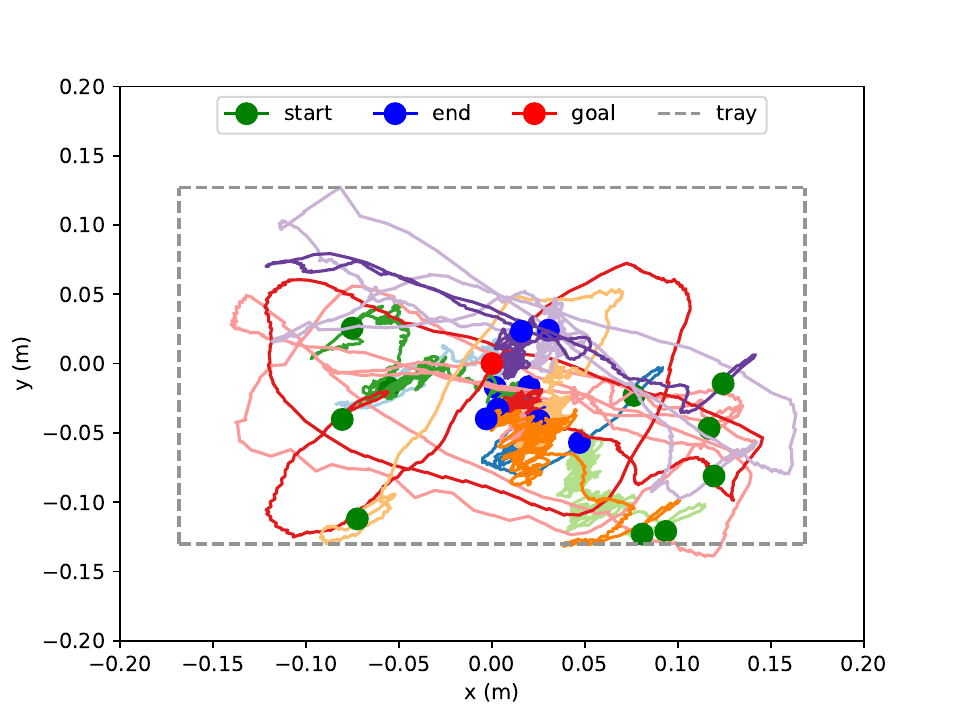}

    \caption{Trajectories of ball in tray frame for 10 different episodes of the dynamic balancing task. At the start of each episode the robot starts from the home configuration and the ball is placed at an arbitrary location on the tray by the human operator. Our control framework is able to achieve a median error of $3.9$ cm. Note that due to perception errors, sometimes the center of the ball is detected to be outside the tray when near the edge.}
    \label{fig:ball_trajectories}
\end{figure}

\textbf{Analysis}: Fig.~\ref{fig:ball_trajectories} shows a superimposed plot of the (x,y) trajectories ball with respect to the tray frame for 10 different trials of the experiment. 
We observe that, even with our highly biased model and noise due to perception and state estimation, our MPC framework is able to achieve a median error of $3.9$ cm in positioning the ball at the center of the tray. 
Moreover, the robot did not drop the ball in any trial. This experiment demonstrates the efficacy of MPC in correcting for model bias while maintaining reactivity and handling complex task constraints. 


Our experimental results indicate that accurate and reactive manipulation behavior can be obtained by using relatively simple models and intuitive cost functions in an MPC framework. Sampling-based MPC also provides us the  flexibility to encode different desired behaviors such as smooth motions directly in the optimization and tightly couple perception with control which are key components for real-world robot control.


\section{Related Work}
\label{sec:related-work}
Perception driven feedback control on high dimensional systems is a large field of research with several existing approaches~\cite{kappler_reactive}. 
Operation Space Controllers~(OSC) are some of the fastest algorithms available for feedback control, with methods achieving control latency of 1-2~ms~\cite{dietrich2012reactive,cheng2018rmpflow,haviland2020purely,Rakita-RSS-18}. However OSC methods rely heavily on a higher level planner for avoiding local minima~(e.g.,obstacles).

A more global approach has been explored by reformulating standard motion planning methods to do feedback control via online replanning~\cite{kuntz2020fast,torres2015motion,karaman2011anytime,alwala2020joint,murray2016robot}. However most of these methods run at a slow rates on high dimensional systems, with control latencies between 140ms and 1000ms~\cite{kuntz2020fast,torres2015motion,karaman2011anytime,alwala2020joint}. Murray~\etal~\cite{murray2016robot} researched leveraging a custom chip to do fast parallel collision checking and use this with a PRM style planner to replan at 1ms for reaching Cartesian Poses in a semi-structured environment. Their chip based collision checker uses a complete pointcloud of the environment obtained by placing many cameras in the environment and combining their pointclouds. This is an highly unrealistic setting for real world manipulation in unstructured environments.

In the realm of Model predictive control approaches, only gradient based joint space MPC methods have shown to work on real manipulation systems as their control latency is in an acceptable range~(20ms - 125ms ) for feedback control~\cite{fishman2020collaborative,erez_humanoidmpc_2013,ishihara2019full}. 
Ishihara~\etal~\cite{ishihara2019full} explore two stage hierarchical ILQR for fast MPC on a humanoid robot. Their two stage approach enables a very low control latency of 20ms. Erez~\etal~\cite{erez_humanoidmpc_2013} explore ILQR on humanoid robots leveraging a simulator for the dynamics model. Fishman~\etal~\cite{fishman2020collaborative} use gradient based MPC for finding trajectories for a manipulator while simultaneously predicting user intent in a human robot interaction setting. They solve the optimization problem leveraging Levenberg-Marquardt at a control latency of~140ms. Hogan and Rodriguez~\cite{hogan2020reactive} explore gradient based MPC in the task space for planar non-prehensile manipulation leveraging a learned mode switcher to switch between different dynamic models. They are able to obtain a control latency of~5ms as their dynamic models are smooth.


Sampling-based methods have a rich history in MPC. Model-Predictive Path Integral Control (MPPI)~\citep{williams2016aggressive} is one of the leading sampling-based MPC approaches that has shown great performance on real-world aggressive off-road autonomous driving by leveraging learned models~\citep{williams2017information} and GPU acceleration~\citep{williams2017model}. 
Wagener~\etal~\cite{wagener19a} analyze MPC algorithms from the perspective of online learning and show connections between different methods such as Cross-Entropy method~(CEM) and MPPI and have also demonstrated control rates of 40Hz with 1200 samples and a horizon of 2.5 seconds for off-road driving using GPU acceleration.

However, in the context of manipulation, 
sampling-based control in joint space has only been explored by optimizing in the joint position space without considering velocity and acceleration limits~\cite{danielczuk2020object,hyatt2020parameterized,hyatt2020real}.  Danielczuk~\etal~\cite{danielczuk2020object} learn a collision classifier and do online replanning in joint space leveraging an inverse kinematic function to get a goal joint configuration and find straight line paths in joint space to reach the goal while avoiding collisions. Their approach doesn't handle different task spaces directly in the form of cost functions and also has a much larger control latency of 1000 ms. Hyatt~\etal~\cite{hyatt2020parameterized,hyatt2020real} compare sampling based MPC with gradient based MPC on large dimensional robots with piecewise linear functions. They show that sampling based MPC can run at 200Hz~(5ms) even with large number of dimensions due to it's parallelizability on the GPU. However, they do not explore collision avoidance or task space cost terms in their approach and leave it for future work. Pinneri~\etal~\cite{pinneri2020sample} provide a method for adding correlated noise to action samples in Cross-Entropy Method to reduce the number of particles required for obtaining good performance on high-dimensional control tasks. However, their multi-threaded CPU implementation is unable to achieve real-time  performance. Their experiments are also limited to simulation with access to true dynamics. 

Sampling based optimization has also been used for motion planning in high dimensional systems~\cite{stomp,kobilarov2012cross,kobilarov2012crossijrr}. Kalakrishnan~\etal~\cite{stomp} formulate planning as a stochastic trajectory optimization problem and plan over the joint position space to reach Cartesian positions while orientation constraints on the end-effector. They structure their co-variance matrix based on the finite difference matrix to sample delta joint positions that start and stop at 0 with a smooth profile in between. This sampling along with projecting the weighted action through this matrix, pushes the optimization towards a smooth trajectory for execution on the real robot. Kobilarov~\cite{kobilarov2012cross,kobilarov2012crossijrr} explored using cross-entropy optimization for motion planning and showed global convergence in very tight environments on quadrotors and simple planar environments.


\vspace{-0.2cm}
\section{Discussion}
\label{sec:discussion}
\vspace{-0.2cm}
We presented a sampling-based MPC framework for manipulation that operates in the joint space and is able achieve smooth and reactive motions while respecting constraints, and demonstrated its performance on dynamic control tasks. The first key component of our approach is a fully tensorized kinematic model that allows for GPU-based acceleration of rollouts. Second, we leverage intuitive cost terms that encourage desirable behaviors. 
Third, our formulation allows us to leverage diverse sampling strategies to embed desirable properties directly into the optimization. 
However, a few key questions remain. First, performance can be made more robust by directly accounting for state uncertainty in the control loop.
Second, at higher speeds, the kinematic model might induce significant model-bias. Here, learning a residual dynamics model~\cite{zeng2020tossingbot} or a terminal Q-function~\citep{bhardwaj2020information} can help mitigate the effects of model-bias while still maintaining computational speed.


\bibliography{references}

\begin{thebibliography}{51}
\providecommand{\natexlab}[1]{#1}
\providecommand{\url}[1]{\texttt{#1}}
\expandafter\ifx\csname urlstyle\endcsname\relax
  \providecommand{\doi}[1]{doi: #1}\else
  \providecommand{\doi}{doi: \begingroup \urlstyle{rm}\Url}\fi

\bibitem[Bohg et~al.(2017)Bohg, Hausman, Sankaran, Brock, Kragic, Schaal, and
  Sukhatme]{bohg2017interactive}
J.~Bohg, K.~Hausman, B.~Sankaran, O.~Brock, D.~Kragic, S.~Schaal, and G.~S.
  Sukhatme.
\newblock Interactive perception: Leveraging action in perception and
  perception in action.
\newblock \emph{IEEE Transactions on Robotics}, 33\penalty0 (6):\penalty0
  1273--1291, 2017.

\bibitem[Kappler et~al.(2018)Kappler, Meier, Issac, Mainprice, Cifuentes,
  W{\"u}thrich, Berenz, Schaal, Ratliff, and Bohg]{kappler2018real}
D.~Kappler, F.~Meier, J.~Issac, J.~Mainprice, C.~G. Cifuentes, M.~W{\"u}thrich,
  V.~Berenz, S.~Schaal, N.~Ratliff, and J.~Bohg.
\newblock Real-time perception meets reactive motion generation.
\newblock \emph{IEEE Robotics and Automation Letters}, 3\penalty0 (3):\penalty0
  1864--1871, 2018.

\bibitem[Luo and Hauser(2017)]{luo2017robust}
J.~Luo and K.~Hauser.
\newblock Robust trajectory optimization under frictional contact with
  iterative learning.
\newblock \emph{Autonomous Robots}, 41\penalty0 (6):\penalty0 1447--1461, 2017.

\bibitem[Nakanishi et~al.(2008)Nakanishi, Cory, Mistry, Peters, and
  Schaal]{nakanishi2008operational}
J.~Nakanishi, R.~Cory, M.~Mistry, J.~Peters, and S.~Schaal.
\newblock Operational space control: A theoretical and empirical comparison.
\newblock \emph{The International Journal of Robotics Research}, 27\penalty0
  (6):\penalty0 737--757, 2008.
\newblock \doi{10.1177/0278364908091463}.
\newblock URL \url{https://doi.org/10.1177/0278364908091463}.

\bibitem[Cheng et~al.(2018)Cheng, Mukadam, Issac, Birchfield, Fox, Boots, and
  Ratliff]{cheng2018rmpflow}
C.-A. Cheng, M.~Mukadam, J.~Issac, S.~Birchfield, D.~Fox, B.~Boots, and
  N.~Ratliff.
\newblock Rmpflow: A computational graph for automatic motion policy
  generation.
\newblock In \emph{International Workshop on the Algorithmic Foundations of
  Robotics}, pages 441--457. Springer, 2018.

\bibitem[Dietrich et~al.(2012)Dietrich, Wimbock, Albu-Schaffer, and
  Hirzinger]{dietrich2012reactive}
A.~Dietrich, T.~Wimbock, A.~Albu-Schaffer, and G.~Hirzinger.
\newblock Reactive whole-body control: Dynamic mobile manipulation using a
  large number of actuated degrees of freedom.
\newblock \emph{IEEE Robotics \& Automation Magazine}, 19\penalty0
  (2):\penalty0 20--33, 2012.

\bibitem[Bangura and Mahony(2014)]{bangura2014real}
M.~Bangura and R.~Mahony.
\newblock Real-time model predictive control for quadrotors.
\newblock \emph{IFAC Proceedings Volumes}, 47\penalty0 (3):\penalty0
  11773--11780, 2014.

\bibitem[{Erez} et~al.(2013){Erez}, {Lowrey}, {Tassa}, {Kumar}, {Kolev}, and
  {Todorov}]{erez_humanoidmpc_2013}
T.~{Erez}, K.~{Lowrey}, Y.~{Tassa}, V.~{Kumar}, S.~{Kolev}, and E.~{Todorov}.
\newblock An integrated system for real-time model predictive control of
  humanoid robots.
\newblock In \emph{2013 13th IEEE-RAS International Conference on Humanoid
  Robots (Humanoids)}, pages 292--299, 2013.
\newblock \doi{10.1109/HUMANOIDS.2013.7029990}.

\bibitem[{Scianca} et~al.(2020){Scianca}, {De Simone}, {Lanari}, and
  {Oriolo}]{sciana_qphumanoidmpc_2020}
N.~{Scianca}, D.~{De Simone}, L.~{Lanari}, and G.~{Oriolo}.
\newblock Mpc for humanoid gait generation: Stability and feasibility.
\newblock \emph{IEEE Transactions on Robotics}, 36\penalty0 (4):\penalty0
  1171--1188, 2020.
\newblock \doi{10.1109/TRO.2019.2958483}.

\bibitem[Williams et~al.(2016)Williams, Drews, Goldfain, Rehg, and
  Theodorou]{williams2016aggressive}
G.~Williams, P.~Drews, B.~Goldfain, J.~M. Rehg, and E.~A. Theodorou.
\newblock Aggressive driving with model predictive path integral control.
\newblock In \emph{2016 IEEE International Conference on Robotics and
  Automation (ICRA)}, pages 1433--1440. IEEE, 2016.

\bibitem[Tassa et~al.(2014)Tassa, Mansard, and Todorov]{tassa2014control}
Y.~Tassa, N.~Mansard, and E.~Todorov.
\newblock Control-limited differential dynamic programming.
\newblock In \emph{2014 IEEE International Conference on Robotics and
  Automation (ICRA)}, pages 1168--1175. IEEE, 2014.

\bibitem[Williams et~al.(2017{\natexlab{a}})Williams, Aldrich, and
  Theodorou]{williams2017model}
G.~Williams, A.~Aldrich, and E.~A. Theodorou.
\newblock Model predictive path integral control: From theory to parallel
  computation.
\newblock \emph{Journal of Guidance, Control, and Dynamics}, 40\penalty0
  (2):\penalty0 344--357, 2017{\natexlab{a}}.

\bibitem[Williams et~al.(2017{\natexlab{b}})Williams, Wagener, Goldfain, Drews,
  Rehg, Boots, and Theodorou]{williams2017information}
G.~Williams, N.~Wagener, B.~Goldfain, P.~Drews, J.~M. Rehg, B.~Boots, and E.~A.
  Theodorou.
\newblock Information theoretic mpc for model-based reinforcement learning.
\newblock In \emph{2017 IEEE International Conference on Robotics and
  Automation (ICRA)}, pages 1714--1721. IEEE, 2017{\natexlab{b}}.

\bibitem[Wagener et~al.(2019)Wagener, Cheng, Sacks, and Boots]{wagener19a}
N.~Wagener, C.-A. Cheng, J.~Sacks, and B.~Boots.
\newblock An online learning approach to model predictive control.
\newblock 2019.

\bibitem[Chua et~al.(2018)Chua, Calandra, McAllister, and Levine]{chua2018deep}
K.~Chua, R.~Calandra, R.~McAllister, and S.~Levine.
\newblock Deep reinforcement learning in a handful of trials using
  probabilistic dynamics models.
\newblock \emph{arXiv preprint arXiv:1805.12114}, 2018.

\bibitem[Nagabandi et~al.(2020)Nagabandi, Konolige, Levine, and
  Kumar]{nagabandi2020deep}
A.~Nagabandi, K.~Konolige, S.~Levine, and V.~Kumar.
\newblock Deep dynamics models for learning dexterous manipulation.
\newblock In \emph{Conference on Robot Learning}, pages 1101--1112. PMLR, 2020.

\bibitem[Danielczuk et~al.(2020)Danielczuk, Mousavian, Eppner, and
  Fox]{danielczuk2020object}
M.~Danielczuk, A.~Mousavian, C.~Eppner, and D.~Fox.
\newblock Object rearrangement using learned implicit collision functions.
\newblock \emph{arXiv preprint arXiv:2011.10726}, 2020.

\bibitem[{Zhong} et~al.(2013){Zhong}, {Johnson}, {Tassa}, {Erez}, and
  {Todorov}]{6614995}
M.~{Zhong}, M.~{Johnson}, Y.~{Tassa}, T.~{Erez}, and E.~{Todorov}.
\newblock Value function approximation and model predictive control.
\newblock In \emph{2013 IEEE Symposium on Adaptive Dynamic Programming and
  Reinforcement Learning (ADPRL)}, pages 100--107, 2013.
\newblock \doi{10.1109/ADPRL.2013.6614995}.

\bibitem[Wagener et~al.(2019)Wagener, Cheng, Sacks, and
  Boots]{wagener2019online}
N.~Wagener, C.-A. Cheng, J.~Sacks, and B.~Boots.
\newblock An online learning approach to model predictive control.
\newblock \emph{arXiv preprint arXiv:1902.08967}, 2019.

\bibitem[Stulp and Sigaud(2012)]{stulp2012path}
F.~Stulp and O.~Sigaud.
\newblock Path integral policy improvement with covariance matrix adaptation.
\newblock \emph{arXiv preprint arXiv:1206.4621}, 2012.

\bibitem[Klein and Blaho(1987)]{klein1987dexterity}
C.~A. Klein and B.~E. Blaho.
\newblock Dexterity measures for the design and control of kinematically
  redundant manipulators.
\newblock \emph{The international journal of robotics research}, 6\penalty0
  (2):\penalty0 72--83, 1987.

\bibitem[Vahrenkamp et~al.(2012)Vahrenkamp, Asfour, Metta, Sandini, and
  Dillmann]{vahrenkamp2012manipulability}
N.~Vahrenkamp, T.~Asfour, G.~Metta, G.~Sandini, and R.~Dillmann.
\newblock Manipulability analysis.
\newblock In \emph{2012 12th ieee-ras international conference on humanoid
  robots (humanoids 2012)}, pages 568--573. IEEE, 2012.

\bibitem[Rakita et~al.(2018)Rakita, Mutlu, and Gleicher]{Rakita-RSS-18}
D.~Rakita, B.~Mutlu, and M.~Gleicher.
\newblock {RelaxedIK: Real-time Synthesis of Accurate and Feasible Robot Arm
  Motion}.
\newblock In \emph{Proceedings of Robotics: Science and Systems}, Pittsburgh,
  Pennsylvania, June 2018.
\newblock \doi{10.15607/RSS.2018.XIV.043}.

\bibitem[Mildenhall et~al.(2020)Mildenhall, Srinivasan, Tancik, Barron,
  Ramamoorthi, and Ng]{mildenhall2020nerf}
B.~Mildenhall, P.~P. Srinivasan, M.~Tancik, J.~T. Barron, R.~Ramamoorthi, and
  R.~Ng.
\newblock Nerf: Representing scenes as neural radiance fields for view
  synthesis.
\newblock In \emph{European Conference on Computer Vision}, pages 405--421.
  Springer, 2020.

\bibitem[Zucker et~al.(2013)Zucker, Ratliff, Dragan, Pivtoraiko, Klingensmith,
  Dellin, Bagnell, and Srinivasa]{zucker2013chomp}
M.~Zucker, N.~Ratliff, A.~D. Dragan, M.~Pivtoraiko, M.~Klingensmith, C.~M.
  Dellin, J.~A. Bagnell, and S.~S. Srinivasa.
\newblock Chomp: Covariant hamiltonian optimization for motion planning.
\newblock \emph{The International Journal of Robotics Research}, 32\penalty0
  (9-10):\penalty0 1164--1193, 2013.

\bibitem[Niederreiter(1992)]{niederreiter1992random}
H.~Niederreiter.
\newblock \emph{Random number generation and quasi-Monte Carlo methods},
  volume~63.
\newblock Society for Industrial and Applied Mathematics, Philadelphia,~US,
  1992.

\bibitem[Halton and Smith(1964)]{Halton64}
J.~H. Halton and G.~B. Smith.
\newblock Algorithm 247: Radical-inverse quasi-random point sequence.
\newblock \emph{Communications {ACM}}, 7\penalty0 (12):\penalty0 701--702,
  1964.

\bibitem[Dick et~al.(2013)Dick, Kuo, and Sloan]{dick2013qmc}
J.~Dick, F.~Y. Kuo, and I.~H. Sloan.
\newblock High-dimensional integration: The quasi-monte carlo way.
\newblock \emph{Acta Numerica}, 22:\penalty0 133--288, 2013.

\bibitem[Kobilarov(2012)]{kobilarov2012cross}
M.~Kobilarov.
\newblock Cross-entropy randomized motion planning.
\newblock In \emph{Robotics: Science and Systems}, volume~7, pages 153--160,
  2012.

\bibitem[Yang et~al.(2020)Yang, Caluwaerts, Iscen, Zhang, Tan, and
  Sindhwani]{yang2020data}
Y.~Yang, K.~Caluwaerts, A.~Iscen, T.~Zhang, J.~Tan, and V.~Sindhwani.
\newblock Data efficient reinforcement learning for legged robots.
\newblock In \emph{Conference on Robot Learning}, pages 1--10. PMLR, 2020.

\bibitem[Summers et~al.(2020)Summers, Lowrey, Rajeswaran, Srinivasa, and
  Todorov]{summers2020lyceum}
C.~Summers, K.~Lowrey, A.~Rajeswaran, S.~Srinivasa, and E.~Todorov.
\newblock Lyceum: An efficient and scalable ecosystem for robot learning.
\newblock In \emph{Learning for Dynamics and Control}, pages 793--803. PMLR,
  2020.

\bibitem[Haviland and Corke(2020)]{haviland2020purely}
J.~Haviland and P.~Corke.
\newblock A purely-reactive manipulability-maximising motion controller.
\newblock \emph{arXiv e-prints}, pages arXiv--2002, 2020.

\bibitem[Ratliff et~al.(2018)Ratliff, Issac, Kappler, Birchfield, and
  Fox]{ratliff2018riemannian}
N.~D. Ratliff, J.~Issac, D.~Kappler, S.~Birchfield, and D.~Fox.
\newblock Riemannian motion policies.
\newblock \emph{arXiv preprint arXiv:1801.02854}, 2018.

\bibitem[{Kappler} et~al.(2018){Kappler}, {Meier}, {Issac}, {Mainprice},
  {Cifuentes}, {Wüthrich}, {Berenz}, {Schaal}, {Ratliff}, and
  {Bohg}]{kappler_reactive}
D.~{Kappler}, F.~{Meier}, J.~{Issac}, J.~{Mainprice}, C.~G. {Cifuentes},
  M.~{Wüthrich}, V.~{Berenz}, S.~{Schaal}, N.~{Ratliff}, and J.~{Bohg}.
\newblock Real-time perception meets reactive motion generation.
\newblock \emph{IEEE Robotics and Automation Letters}, 3\penalty0 (3):\penalty0
  1864--1871, 2018.
\newblock \doi{10.1109/LRA.2018.2795645}.

\bibitem[Kuntz et~al.(2020)Kuntz, Bowen, and Alterovitz]{kuntz2020fast}
A.~Kuntz, C.~Bowen, and R.~Alterovitz.
\newblock Fast anytime motion planning in point clouds by interleaving sampling
  and interior point optimization.
\newblock In \emph{Robotics Research}, pages 929--945. Springer, 2020.

\bibitem[Torres et~al.(2015)Torres, Kuntz, Gilbert, Swaney, Hendrick, Webster,
  and Alterovitz]{torres2015motion}
L.~G. Torres, A.~Kuntz, H.~B. Gilbert, P.~J. Swaney, R.~J. Hendrick, R.~J.
  Webster, and R.~Alterovitz.
\newblock A motion planning approach to automatic obstacle avoidance during
  concentric tube robot teleoperation.
\newblock In \emph{2015 IEEE International Conference on Robotics and
  Automation (ICRA)}, pages 2361--2367. IEEE, 2015.

\bibitem[Karaman et~al.(2011)Karaman, Walter, Perez, Frazzoli, and
  Teller]{karaman2011anytime}
S.~Karaman, M.~R. Walter, A.~Perez, E.~Frazzoli, and S.~Teller.
\newblock Anytime motion planning using the rrt.
\newblock In \emph{2011 IEEE International Conference on Robotics and
  Automation}, pages 1478--1483. IEEE, 2011.

\bibitem[Alwala and Mukadam(2020)]{alwala2020joint}
K.~V. Alwala and M.~Mukadam.
\newblock Joint sampling and trajectory optimization over graphs for online
  motion planning.
\newblock \emph{arXiv preprint arXiv:2011.07171}, 2020.

\bibitem[Murray et~al.(2016)Murray, Floyd-Jones, Qi, Sorin, and
  Konidaris]{murray2016robot}
S.~Murray, W.~Floyd-Jones, Y.~Qi, D.~J. Sorin, and G.~D. Konidaris.
\newblock Robot motion planning on a chip.
\newblock In \emph{Robotics: Science and Systems}, 2016.

\bibitem[{Fishman} et~al.(2019){Fishman}, {Paxton}, {Yang}, {Fox}, {Boots}, and
  {Ratliff}]{fishman2020collaborative}
A.~{Fishman}, C.~{Paxton}, W.~{Yang}, D.~{Fox}, B.~{Boots}, and N.~{Ratliff}.
\newblock {Collaborative Behavior Models for Optimized Human-Robot Teamwork}.
\newblock \emph{arXiv e-prints}, art. arXiv:1910.04339, Oct. 2019.

\bibitem[Ishihara et~al.(2019)Ishihara, Itoh, and Morimoto]{ishihara2019full}
K.~Ishihara, T.~D. Itoh, and J.~Morimoto.
\newblock Full-body optimal control toward versatile and agile behaviors in a
  humanoid robot.
\newblock \emph{IEEE Robotics and Automation Letters}, 5\penalty0 (1):\penalty0
  119--126, 2019.

\bibitem[Hogan and Rodriguez(2020)]{hogan2020reactive}
F.~R. Hogan and A.~Rodriguez.
\newblock Reactive planar non-prehensile manipulation with hybrid model
  predictive control.
\newblock \emph{The International Journal of Robotics Research}, 39\penalty0
  (7):\penalty0 755--773, 2020.

\bibitem[Hyatt et~al.(2020)Hyatt, Williams, and
  Killpack]{hyatt2020parameterized}
P.~Hyatt, C.~S. Williams, and M.~D. Killpack.
\newblock Parameterized and gpu-parallelized real-time model predictive control
  for high degree of freedom robots.
\newblock \emph{arXiv preprint arXiv:2001.04931}, 2020.

\bibitem[Hyatt and Killpack(2020)]{hyatt2020real}
P.~Hyatt and M.~D. Killpack.
\newblock Real-time nonlinear model predictive control of robots using a
  graphics processing unit.
\newblock \emph{IEEE Robotics and Automation Letters}, 5\penalty0 (2):\penalty0
  1468--1475, 2020.

\bibitem[Pinneri et~al.(2020)Pinneri, Sawant, Blaes, Achterhold, Stueckler,
  Rolinek, and Martius]{pinneri2020sample}
C.~Pinneri, S.~Sawant, S.~Blaes, J.~Achterhold, J.~Stueckler, M.~Rolinek, and
  G.~Martius.
\newblock Sample-efficient cross-entropy method for real-time planning.
\newblock \emph{arXiv preprint arXiv:2008.06389}, 2020.

\bibitem[{Kalakrishnan} et~al.(2011){Kalakrishnan}, {Chitta}, {Theodorou},
  {Pastor}, and {Schaal}]{stomp}
M.~{Kalakrishnan}, S.~{Chitta}, E.~{Theodorou}, P.~{Pastor}, and S.~{Schaal}.
\newblock Stomp: Stochastic trajectory optimization for motion planning.
\newblock In \emph{2011 IEEE International Conference on Robotics and
  Automation}, pages 4569--4574, 2011.
\newblock \doi{10.1109/ICRA.2011.5980280}.

\bibitem[Kobilarov(2012)]{kobilarov2012crossijrr}
M.~Kobilarov.
\newblock Cross-entropy motion planning.
\newblock \emph{The International Journal of Robotics Research}, 31\penalty0
  (7):\penalty0 855--871, 2012.

\bibitem[Zeng et~al.(2020)Zeng, Song, Lee, Rodriguez, and
  Funkhouser]{zeng2020tossingbot}
A.~Zeng, S.~Song, J.~Lee, A.~Rodriguez, and T.~Funkhouser.
\newblock Tossingbot: Learning to throw arbitrary objects with residual
  physics.
\newblock \emph{IEEE Transactions on Robotics}, 36\penalty0 (4):\penalty0
  1307--1319, 2020.

\bibitem[Bhardwaj et~al.(2020)Bhardwaj, Handa, Fox, and
  Boots]{bhardwaj2020information}
M.~Bhardwaj, A.~Handa, D.~Fox, and B.~Boots.
\newblock Information theoretic model predictive q-learning.
\newblock In \emph{Learning for Dynamics and Control}, pages 840--850. PMLR,
  2020.

\bibitem[Paszke et~al.(2019)Paszke, Gross, Massa, Lerer, Bradbury, Chanan,
  Killeen, Lin, Gimelshein, Antiga, Desmaison, Kopf, Yang, DeVito, Raison,
  Tejani, Chilamkurthy, Steiner, Fang, Bai, and Chintala]{NEURIPS2019_9015}
A.~Paszke, S.~Gross, F.~Massa, A.~Lerer, J.~Bradbury, G.~Chanan, T.~Killeen,
  Z.~Lin, N.~Gimelshein, L.~Antiga, A.~Desmaison, A.~Kopf, E.~Yang, Z.~DeVito,
  M.~Raison, A.~Tejani, S.~Chilamkurthy, B.~Steiner, L.~Fang, J.~Bai, and
  S.~Chintala.
\newblock Pytorch: An imperative style, high-performance deep learning library.
\newblock In H.~Wallach, H.~Larochelle, A.~Beygelzimer, F.~d\textquotesingle
  Alch\'{e}-Buc, E.~Fox, and R.~Garnett, editors, \emph{Advances in Neural
  Information Processing Systems 32}, pages 8024--8035. 2019.

\bibitem[Sutanto et~al.(2020)Sutanto, Wang, Lin, Mukadam, Sukhatme, Rai, and
  Meier]{sutanto20a}
G.~Sutanto, A.~Wang, Y.~Lin, M.~Mukadam, G.~Sukhatme, A.~Rai, and F.~Meier.
\newblock Encoding physical constraints in differentiable newton-euler
  algorithm.
\newblock volume 120 of \emph{Proceedings of Machine Learning Research}, pages
  804--813, The Cloud, 10--11 Jun 2020. PMLR.
\newblock URL \url{http://proceedings.mlr.press/v120/sutanto20a.html}.

\end{thebibliography}
\newpage
\appendix
\section*{Appendix}

\section{Real-Time Control Implementation}
\label{sec:implementation}

We implemented our MPC pipeline using PyTorch~\citep{NEURIPS2019_9015} with manipulator forward kinematics adapted from the open-source implementation provided by Sutanta~\etal~\citep{sutanto20a} 
and all cost terms and update equations implemented in a batched fashion. Further,  multiprocessing is used to run MPC in a seperate process to avoid latency issues. Table~\ref{tab:mpc_timing_benchmark} shows a timing comparison of our system running on a Titan RTX GPU against leading manipulator control approaches in literature.

\subsection{Franka Panda Control System}
\label{subsec:franka_control}

\begin{figure}[htpb]
\includegraphics[trim=0 300 0 100, clip, width=\linewidth]{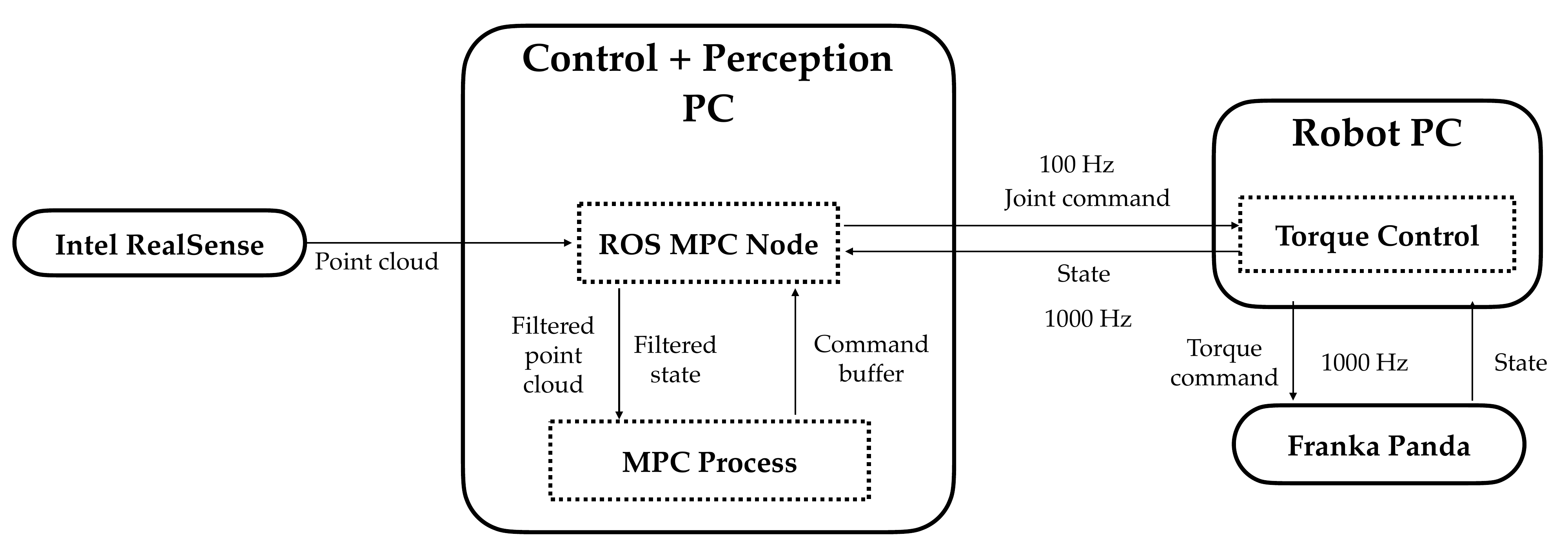}
    \caption{The compute graph shows the flow of information between the different components in our approach.}
    \label{fig:mpc_flowchart}
\end{figure}

The desired acceleration $\ddot{\theta}_t^d$ command from MPC is evaulated at 100Hz and integrated forward to obtain desired joint position ($\theta_t^d$) and joint velocity $\dot{\theta}_t^d$ commands respectively. These commands are sent to a custom low-level torque controller that computes desired torque commands at 1000Hz to control the Franka robot,
\begin{equation}
\label{eq:tau_des_robot}
\tau_t^{ff} = M(\theta_t)\ddot{\theta}_t^d + C(\theta_t)\dot{\theta}_t + K_{p}(\theta_{t}^{d} - \theta_t) + K_{d}(\dot{\theta}_{t}^{d} - \theta_{t})
\end{equation}
where $M(\theta)$ and $C(\dot{\theta})$ are the inertia and coriolis force matrices respectively provided by $\mathtt{libfranka}$ and $K_{p}, K_{d}$ are gains for the position and velocity errors respectively. Fig.~\ref{fig:mpc_flowchart} shows the overall architecture of the control system.

\subsection{State Estimation and Perception}
We found the noise in the joint state read from $\mathtt{libfranka}$, especially in joint velocities, to be prohibitive for precise control with MPC. In order to circumvent this issue we implemented a joint state filter that first predicts the state based on the previous commanded joint acceleration and then uses an exponential moving average filter to incorporate sensor readings.

For our perception setup, we use an Intel Realsense D455 Depth Camera placed at a fixed location in the workspace with a known camera-to-robot transform to obtain depth data in the form of point clouds. The raw point cloud is filtered to remove all the points lying inside the robot body to obtain the \textit{scene} point cloud. The robot URDF is used to sample points along the robot body to create a \textit{robot} point cloud. Both these point clouds are fed as input to SceneCollisionNet~\citep{danielczuk2020object} for computing the collision cost in Eq.~\ref{eq:coll_cost}. Since we only consider static scenes for the purposes of this work, the \textit{scene} pointcloud is computed once at the start of the run. For dynamic scenes, the pointcloud would need to be processed in real-time. We defer this to future work.



\begin{table}
    \centering
    \small
        \caption{Control Latency of methods that can handle collision avoidance in high dimensional systems is tabulated here. A more thorough description is available in Sec.~\ref{sec:related-work}.}
    \label{tab:mpc_timing_benchmark}
    \begin{tabular}{lrr} 
    \toprule
         \textbf{Method} & \textbf{Latency (ms)} &\textbf{Horizon}\\ \toprule
         OSC~\cite{dietrich2012reactive,cheng2018rmpflow,haviland2020purely,Rakita-RSS-18} & 1-17 & 1 \\
         Motion Planning~\cite{torres2015motion,karaman2011anytime,alwala2020joint} & 140-1000 & N/A  \\
         Custom Chip Motion Planning~\cite{murray2016robot} & 1 & N/A  \\
         Gradient MPC~\cite{fishman2020collaborative,erez_humanoidmpc_2013,ishihara2019full} & 20 - 140 & $\leq$16\\
         Sampling MPC~\cite{danielczuk2020object} & 100 & 40 \\ \midrule
         Ours & 10 & 30 \\ \bottomrule
    \end{tabular}
\end{table}

\section{Further Experimental Details}

\subsection{Dynamic Object Balancing}
\label{sec:appendix:ball_balancing}


\textbf{Experimental Setup}: In this task, the real Franka Panda robot is required to balance a ball on tray grasped by the parallel jaw gripper. Every episode starts with the ball placed at an arbitrary location on the tray with the robot trying to center the ball without dropping it. Each episode lasts for 30 seconds after which the ball is placed at an arbitrary location by the human user. The position of the ball is measured at 30Hz using perception system that uses the RGBD input from a RealSense camera. The ball is detected from the RGB images using a blob-tracker and the corresponding depth is queried from the aligned depth image. The camera intrinsics are then used to compute the 3D coordinates. 

\textbf{Ball Dynamics}: We use a simple kinematic model of rolling on plane under acceleration due to gravity to predict the future positions and velocities of the ball in the frame of the tray given the end effector positions and orientations obtained from our arm model. In this simplified model, we do not explicitly account for friction or perform any system identification. In order to predict the future states of the ball given accelerations, we can leverage our tensorized forward model and maintain a high control frequency.  
Let the state of the ball in world frame (as input by the perception system) be denoted by
\begin{align*}
    &x_{ball}^{w} = [x, y, z] &\dot{x}_{ball}^{w} = [\dot{x}, \dot{y}, \dot{z}]
\end{align*}
Further, let the gravity vector in world frame be denoted by $g^{w}$. From forward kinematics, we can calculate the homogeneous transform of the end effector with respect to world frame as 

\begin{equation*}
 T_{ee}^{w} =  \left(\begin{array}{@{}c|c@{}}
    R_{ee}^{w}  & d_{ee}^w \\\hline
    0  & 1 \\
  \end{array}\right)
\end{equation*}

Given a batch of such end-effector poses (obtained from our tensorized arm model rollouts), we can calculate the accelerations due to gravity of the ball in the end effector frame as $g^{ee} = R_{ee}^{w T}g^{w}$ and $\ddot{x}_{obj}^{ee} = \left[g^{ee}_{x}, g^{ee}_{y}, 0\right]$. Here, we made the assumption that the ball does not lose contact with the plate which is reasonable at lower speeds. Similarly, the position and velocity of the ball in end effector frame are calculated as $x_{ball}^{ee} = T_{ee}^{w -1} * x_{ball}^{w}$ and $\dot{x}_{ball}^{ee} = R_{ee}^{w T} * \dot{x}_{ball}^{w}$ with $\dot{x}_{ball}^{ee}[2] = 0$. Finally, given the initial state and a batch of acceleration inputs in the end effector frame, we can simply employ our tensorized kinematic model to predict the future states of the ball.

\subsection{Reaching Cartesian Poses}
\label{sec:appendix:reach_pose_exp}

We test the accuracy and path length of Cartesian pose reaching by selecting a sequence of six hard poses for the robot to reach.
We compare against baseline sampling based planners \rrtconnect and \rrtstar using \moveit and, Manipulability Motion Control~\citep{haviland2020purely}, an operation space controller. In these experiments, the \moveit and \mmc baselines use the position and velocity controllers provided by \texttt{franka\_ros} and \texttt{libfranka} respectively which have been extensively tuned and verified for accuracy. Hence, \mmc represents the best accuracy achievable by a feedback controller.  

\begin{wrapfigure}{r}{0.35\textwidth}
  \centering
  \vspace{-17mm}
  \includegraphics[width=\linewidth]{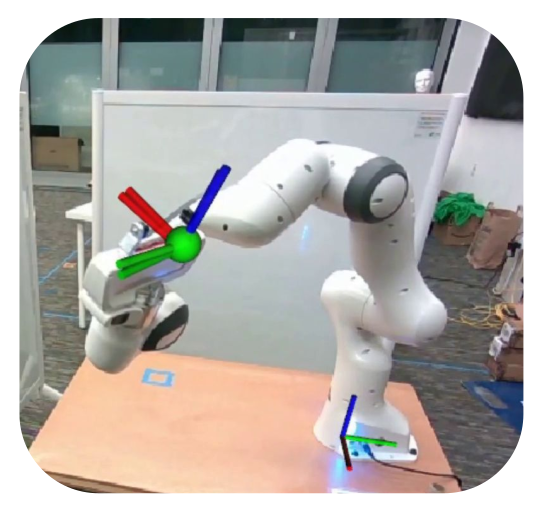}
  \caption{We show the robot trying to reach a very hard orientation (Appendix.~\ref{sec:appendix:reach_pose_exp}). \mmc is unable to reach this pose and continuously enters states of self collision due to its inability to account for collision avoidance while both \mppi and \moveit are able to reach it.}
  \label{fig:pose_reaching_real}
\end{wrapfigure}

One of these poses requires a large change in the end-effector's orientation as shown in Fig.~\ref{fig:pose_reaching_real}. We found that \mmc was unable to handle this pose and would repeatedly result in self collisions owing to its local nature and no consideration of self-collision avoidance, whereas both \storm and \moveit baselines are able to reach the it.
For the other poses, we found the accuracy of \storm to be worse than the baselines (shown in Table~\ref{tab:pose_reaching_results}) although the path lengths and max joint velocities are comparable. Even though our method obtains millimeter level accuracy, the performance is limited by the lower level controller which was not tuned extensively. As future work, we intend to further tune and improve the lower level controller to overcome these issues.

\begin{table}
    \centering
    \footnotesize
    \caption{Comparison of \storm with baselines on pose reaching. Reported values are median across 5 poses excluding the first pose which was unreachable by \mmc. Max Joint Velocity is the maximum velocity achieved throughout the run.}
    \label{tab:pose_reaching_results}
    \begin{tabular}{lrrrrr} 
    \toprule
         \textbf{Method} & \textbf{Pos. Error } &\textbf{Quat. Error} & \textbf{Joint Path Length} & \textbf{EE Path Length } & \textbf{Max. Joint Vel } \\ 
         & (mm) & (\%) & (rad) & (m) & (rad/s) \\\toprule
         \rrtconnect & 0.5328 & 0.054 &3.4527 &1.0086&1.3547\\
         \rrtstar & 0.3433 & 0.0393 &1.7967 &0.7626& 0.7626\\
        \mmc & 0.0661 & 0.00267923 & 2.3394 & 0.74351317&0.268\\
         Ours & 4.822 & 0.4619 & 3.41792 & 1.087688&1.2701\\ \bottomrule
    \end{tabular}

\end{table}

\subsection{SceneCollisionNet Training}\label{sec:appendix:scene_collisionnet}
For the current experiments, we used a pre-trained SceneCollisionnet model~(\href{https://github.com/NVlabs/SceneCollisionNet}{https://github.com/NVlabs/SceneCollisionNet}) provided by the authors of~\citep{danielczuk2020object}. The model was trained for table-top environments which is appropriate for our setting as well.

\section{Ablation Studies}\label{sec:appendix:ablation_studies}
We performed extensive quantitative ablation studies for individual components of our system for the reaching task in simulation. We also present qualitative demonstrations of dynamic obstacle avoidance, effect of horizon and a comparison to Reimannian Motion Policies (RMPs)~\citep{ratliff2018riemannian} on our website \href{https://sites.google.com/view/manipulation-mpc/further-experimental-results}{https://sites.google.com/view/manipulation-mpc/further-experimental-results}.

\subsection{Effect of Number of Particles}
First, we study the effect of number of trajectories sampled per iteration of optimization (or particles) on the controller performance in simulation. For this experiment, 10 end-effector pose targets were used that require significant changes in orientation. Each episode is 700 timesteps long after which the manipulator is reset to the base orientation. The horizon is kept constant at 30 timesteps and all cost function weights are also fixed. 

\subsubsection{Position Accuracy}
The box plot in Fig.~\ref{fig:particles_pos_error} shows the median (solid line) with confidence interval (box) of position errors in the last 50 timesteps of every episode as a function of changing number of particles. We chose the last 50 timesteps to show the convergence of the controller to the goal. From the plot it can be seen that, increasing number of particles the controller is able to achieve more accurate median error with a tighter confidence interval.

\subsubsection{Orientation Accuracy}
The box plot in Fig.~\ref{fig:particles_quat_error} shows the median (solid line) with confidence interval (box) of quaternion errors in the last 50 timesteps of every episode as a function of changing number of particles. Our framework achieves a confidence interval over quaternion errors within 5

\subsubsection{Jerk}
The box plot in Fig.~\ref{fig:particles_jerk} shows the median (solid line) with confidence interval (box) of the jerk in the robot motion over all the timesteps and episodes. Our sampling strategy generates smooth (low jerk) motions even with 200 particles.

\subsubsection{Maximum Joint Velocity}
\begin{figure}
    \centering
    \begin{subfigure}[b]{0.45\textwidth}
        \centering
        \includegraphics[width=\textwidth]{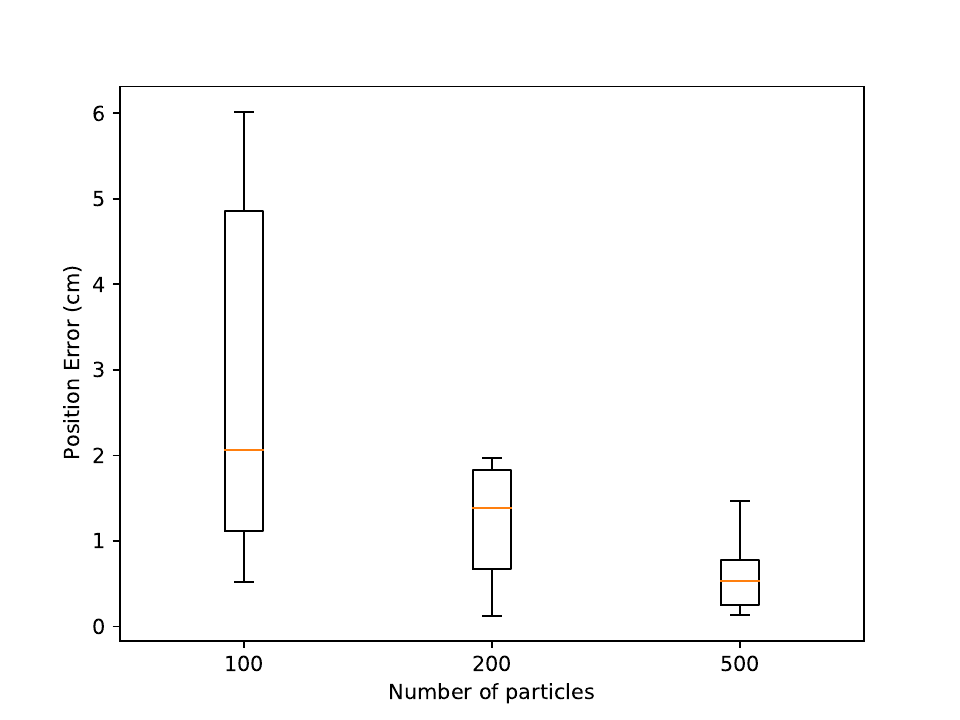}
        \caption{Median position error with confidence bounds as a function of number of particles. }
        \label{fig:particles_pos_error}
    \end{subfigure}
    \hfill
    \begin{subfigure}[b]{0.45\textwidth}
        \centering
        \includegraphics[width=\textwidth]{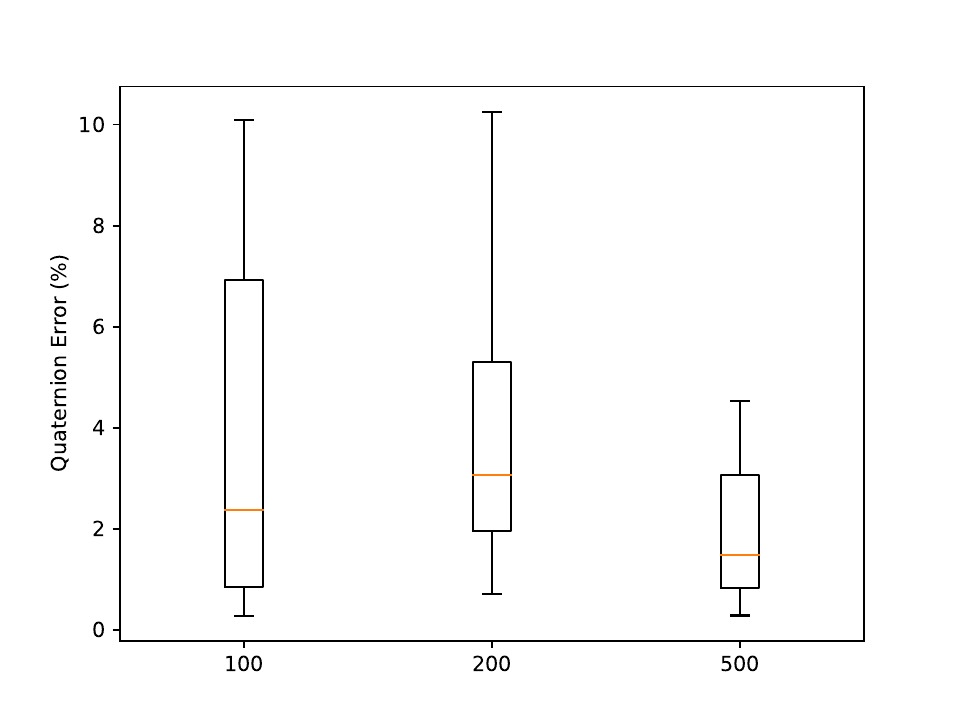}
        \caption{Median quaternion error with confidence bounds as a function of number of particles.}
        \label{fig:particles_quat_error}
    \end{subfigure}
    \hfill
    \begin{subfigure}[b]{0.45\textwidth}
        \centering
        \includegraphics[width=\textwidth]{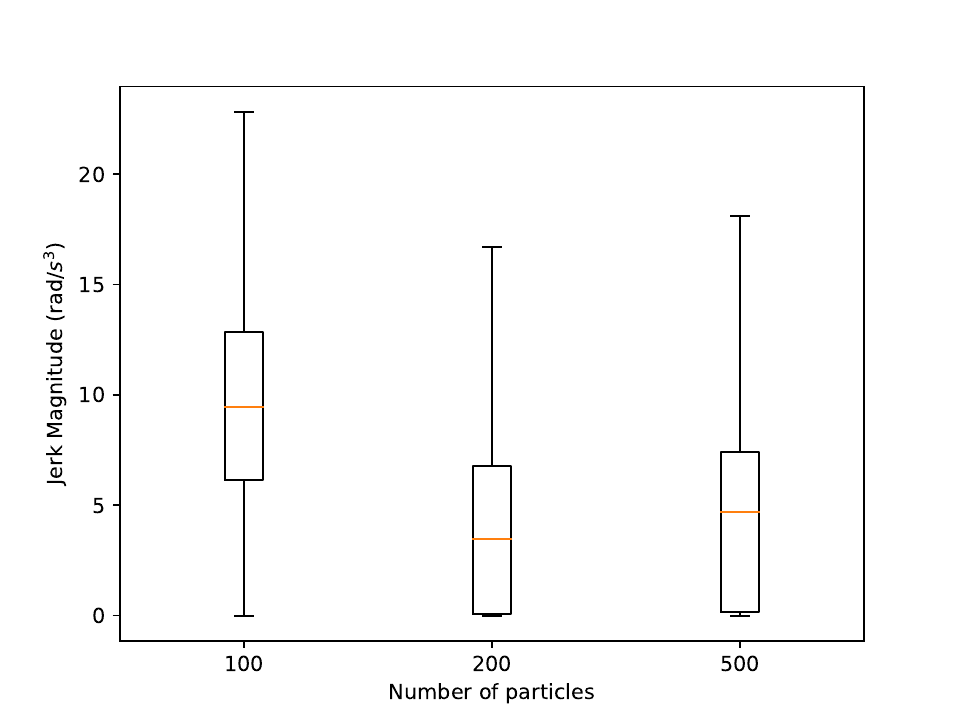}
        \caption{Jerk in robot motion as a function of number of particles.}
        \label{fig:particles_jerk}
    \end{subfigure}
        \begin{subfigure}[b]{0.45\textwidth}
        \centering
        \includegraphics[width=\textwidth]{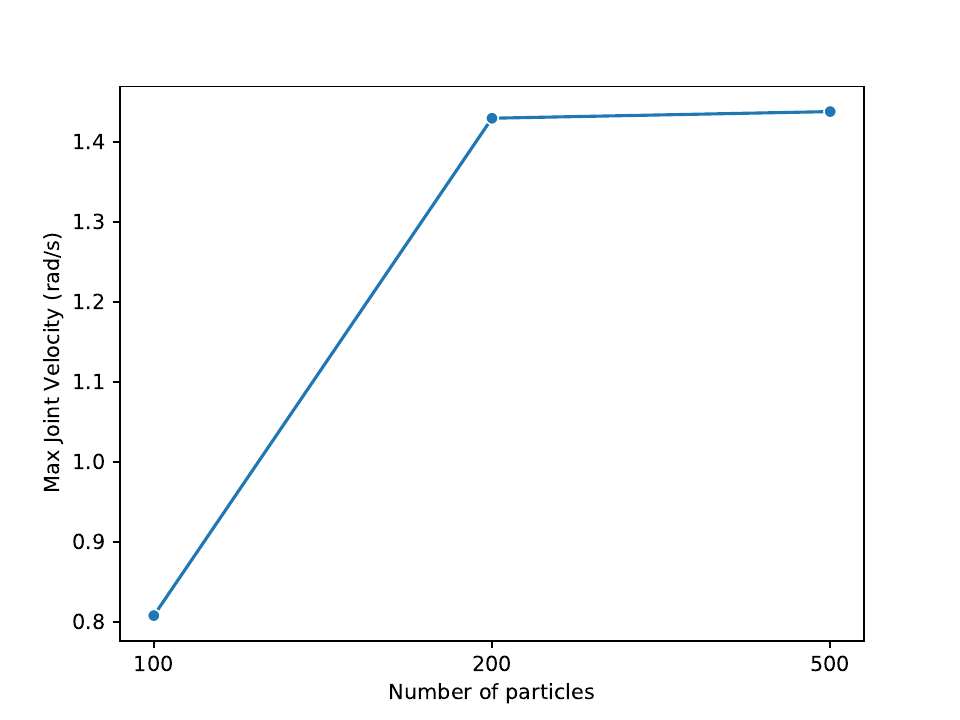}
        \caption{Maximum joint velocity reached for different number of particles.}
        \label{fig:particles_max_vel}
    \end{subfigure}
    \hfill
    \caption{Results for ablation study for number of sampled particles in MPPI.}
    \label{fig:number_of_particles}
\end{figure}

Fig.~\ref{fig:particles_max_vel} demostrates that with increasing number of particles, our B-spline sampling strategy is able to ramp up the robot's joint velocity while maintaining low jerk.

\subsection{Cost Terms}
We study the effect of different cost terms on the controller performance. We test the constraint-based self collision and joint limit avoidance terms and  behavior-based manipulability and stop costs.

\subsubsection{Self-Collision Cost}
For this experiment, we chose 5 end effector target poses that are in collision with the robot and report the number of timesteps the robot spent in self collision. Table~3 shows that when the self collision cost is used, the robot never enters a state of self collision at the cost of not reaching the goal pose.

\subsubsection{Joint Limit Avoidance Cost}
For testing joint limit avoidance 10 end-effector targets with 500 particles were used with a varying weight on the joint limit cost. Table~4 shows that as the weight on the joint limit avoidance cost is increased, the number of violations steadily decreases, going to zero for a large weight of 500 and above.

\begin{table}[]
\centering
\footnotesize
\parbox{.45\linewidth}{
\begin{tabular}{rr}
\toprule
\textbf{Weight} & \textbf{Self Collision (count)} \\ \toprule
0               & 2730                                       \\ 
50              & 0                                          \\ 
500             & 0                                          \\ 
1000            & 0                                          \\ 
5000            & 0                                          \\ 
\bottomrule
\end{tabular}
\vspace{1mm}
\caption{Number of timesteps spent in self collision for changing weight on self collision cost}
}
\hfill
\parbox{.45\linewidth}{
\begin{tabular}{rr}
\toprule
\textbf{Weight} & \textbf{Joint Limits Violation (count)} \\ \toprule
0               & 1624                                       \\ 
50              & 218                                        \\ 
100             & 49                                         \\ 
500             & 0                                          \\ 
1000            & 0                                          \\ \bottomrule
\end{tabular}
\vspace{1mm}
\caption{Number of timesteps spent in joint limit violation for changing weight on joint limit cost}
}

\end{table}

\subsubsection{Manipulability Cost}
The manipulability cost acts as a regularizer to keep the manipulator away from singular configurations. The box plots in Fig.~\ref{fig:manip_pos_error} and Fig.~\ref{fig:manip_quat_error} show the median value with confidence bounds of the position and orientation errors over the last 50 timesteps in 10 different pose reaching runs. We chose the last 50 timesteps and not just the last timestep to test the convergence to the goal.  
Here we see that as the weight on the manipulability cost is increased, the pose reaching accuracy improves. However, after a certain threshold, the manipulability cost interferes with pose reaching and the position accuracy decreases. Maintaining high manipulability allows the robot to reach different end effector orientations accurately.

\subsubsection{Stop Cost}
\begin{figure}
    \centering
    \begin{subfigure}[b]{0.32\textwidth}
        \centering
        \includegraphics[width=\textwidth]{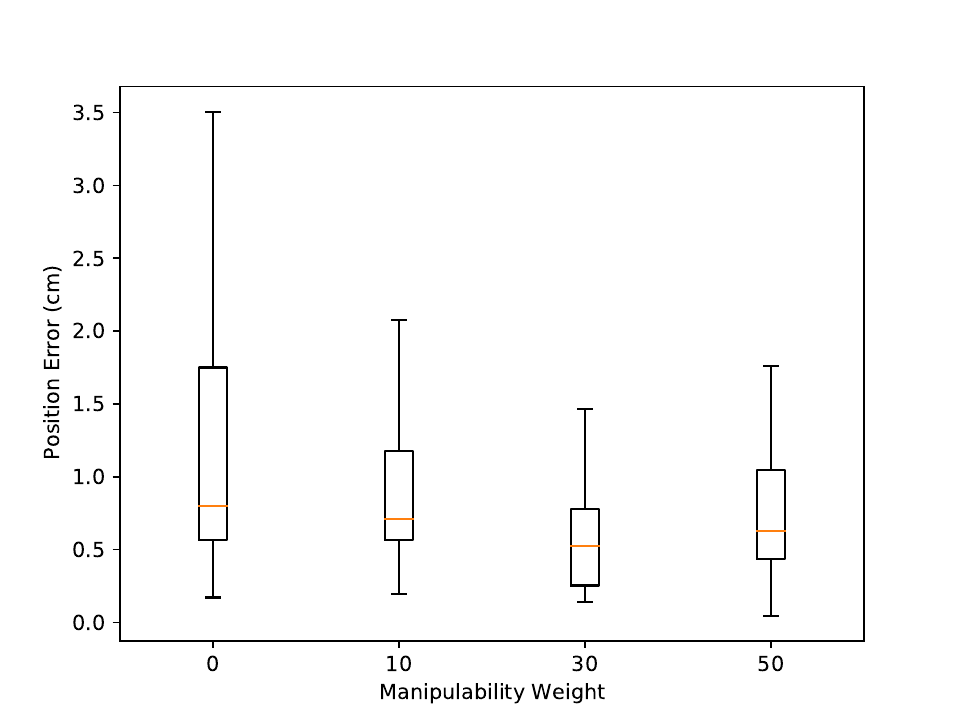}
        \caption{Median position error with confidence bounds as a function of manipulability cost weight. }
        \label{fig:manip_pos_error}
    \end{subfigure}
    \hfill
    \begin{subfigure}[b]{0.32\textwidth}
        \centering
        \includegraphics[width=\textwidth]{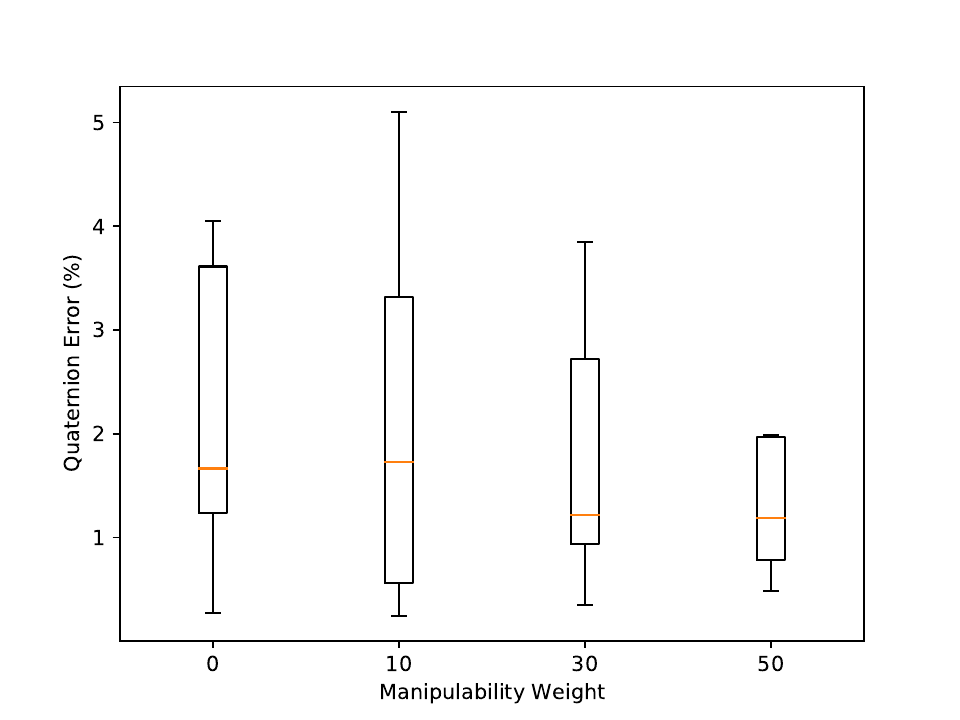}
        \caption{Median quaternion error with confidence bounds as a function of manipulability cost weight.}
        \label{fig:manip_quat_error}
    \end{subfigure}
    \hfill
    \begin{subfigure}[b]{0.32\textwidth}
        \centering
        \includegraphics[width=\textwidth]{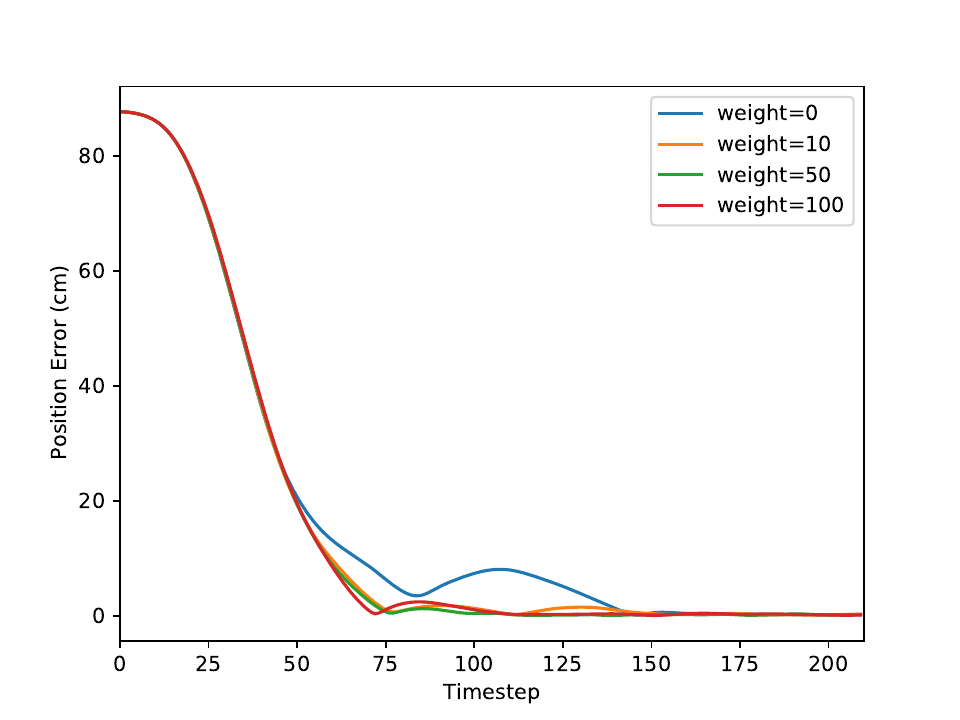}
        \caption{Position error over time for changing stop cost weight. Higher weight avoids oscillation near goal.}
        \label{fig:stop_cost}
    \end{subfigure}
    \hfill
    \caption{Results for ablation study for behavior based manipulability and stop costs.}
    \label{fig:manipulability_and_stop_cost}
\end{figure}

The stop cost penalizes joint velocities that are too high for the robot to safely stop within horizon based on a maximum acceleration threshold. An important consequence of this term is that it allows the robot to smoothly stop at the goal even with a short horizon. We demonstrate this effect in the plot in Fig.~\ref{fig:stop_cost} where a lower weight on the stop cost leads to undesirable oscillations near the goal.

\subsection{Sampling Strategy}
\subsubsection{Pseudo-random vs Halton Sampling}
\begin{figure}
    \centering
    \begin{subfigure}[b]{0.45\textwidth}
        \centering
        \includegraphics[width=\textwidth]{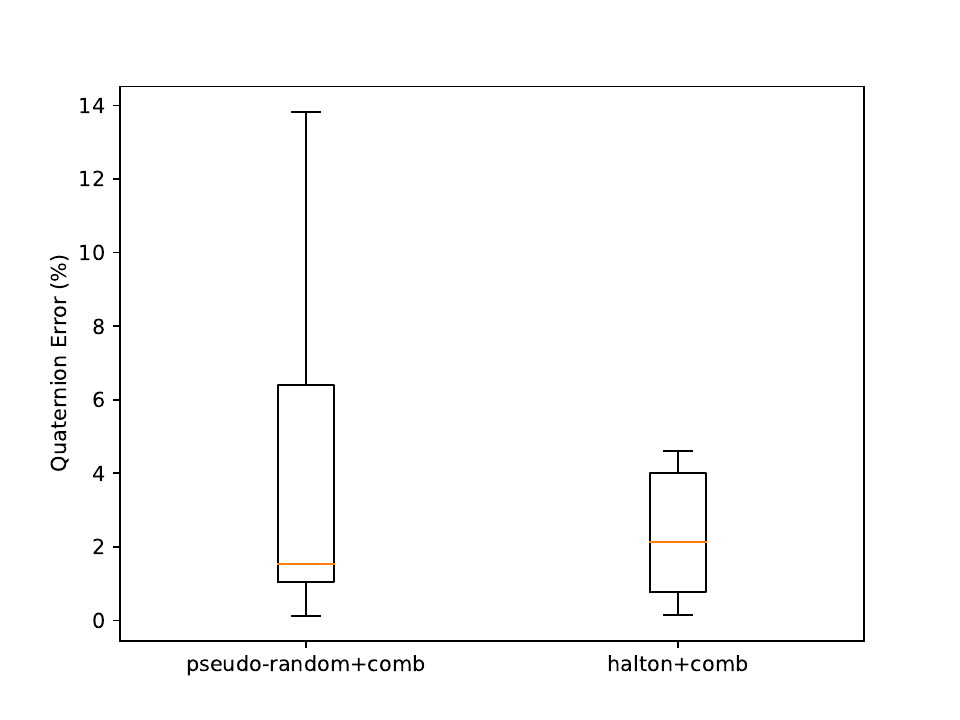}
        \caption{100 particles . }
        \label{fig:pseudo_random_halton_100}
    \end{subfigure}
    \hfill
    \begin{subfigure}[b]{0.45\textwidth}
        \centering
        \includegraphics[width=\textwidth]{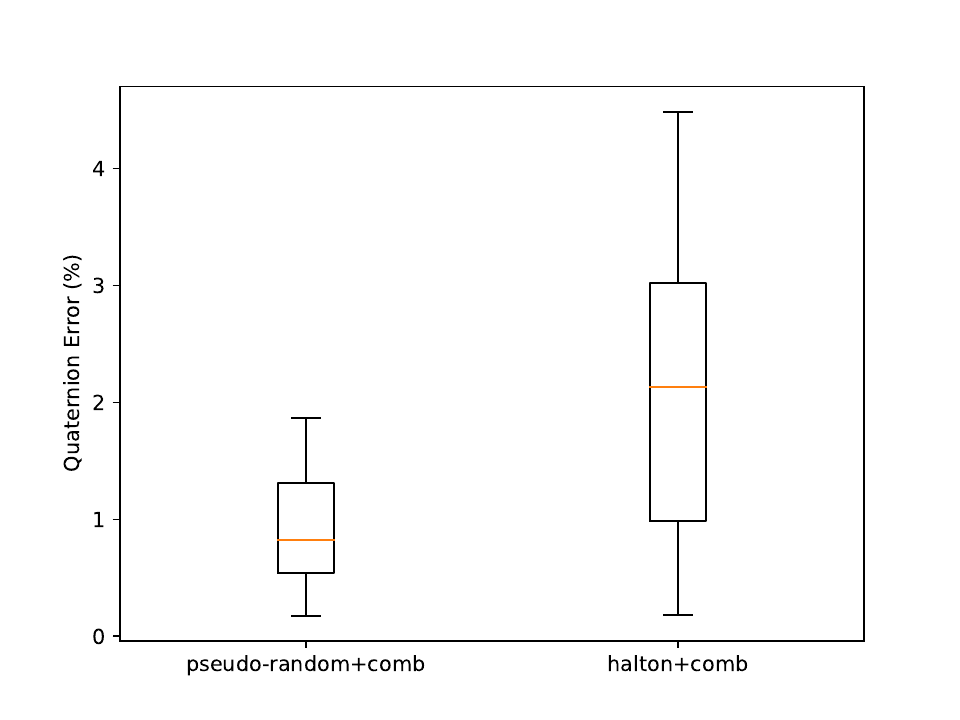}
        \caption{500 particles}
        \label{fig:pseudo_random_halton_500}
    \end{subfigure}
    \caption{Median quaternion error with confidence bounds for psedurandom and Halton sampling with 100 and 500 particles}
    \label{fig:pseudo_random_vs_halton}
\end{figure}

Halton Sampling is a Quasi Monte-Carlo method that provides better coverage of the action space as compared to pseudo-random sampling that exhibits and undesired clustering of samples. We study the improvement gain by using Halton sampling in a low particle regime. The plots in Fig.~\ref{fig:pseudo_random_vs_halton} show the quaternion errors achieved by pseudo random and Halton samples with 100 and 500 particles respectively. With lesser number of particles, pseudo-random sampling with comb filter is unable to achieve a quaternion error with a confidence interval within $5\%$ whereas Halton sampling achieves less than $5\%$ quaternion error with both 100 and 500 particles.

\begin{figure}
    \centering
    \begin{subfigure}[b]{0.32\textwidth}
        \centering
\includegraphics[width=\textwidth]{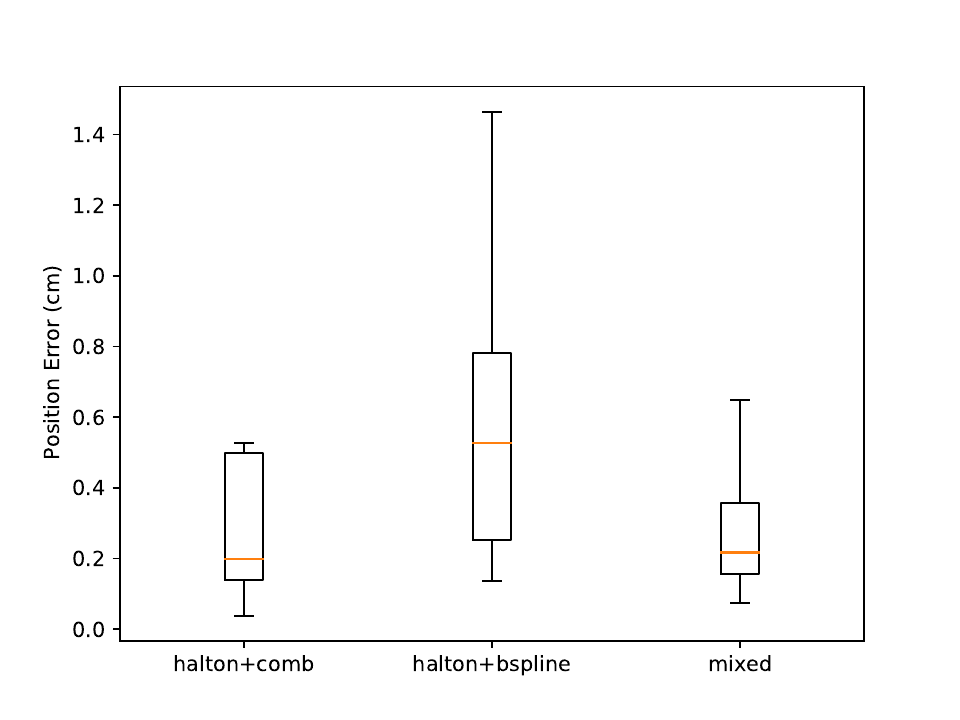}
        \caption{Position Error}
        \label{fig:comb_bspline_mixed_pos_error}
    \end{subfigure}
    \hfill
    \begin{subfigure}[b]{0.32\textwidth}
        \centering
        \includegraphics[width=\textwidth]{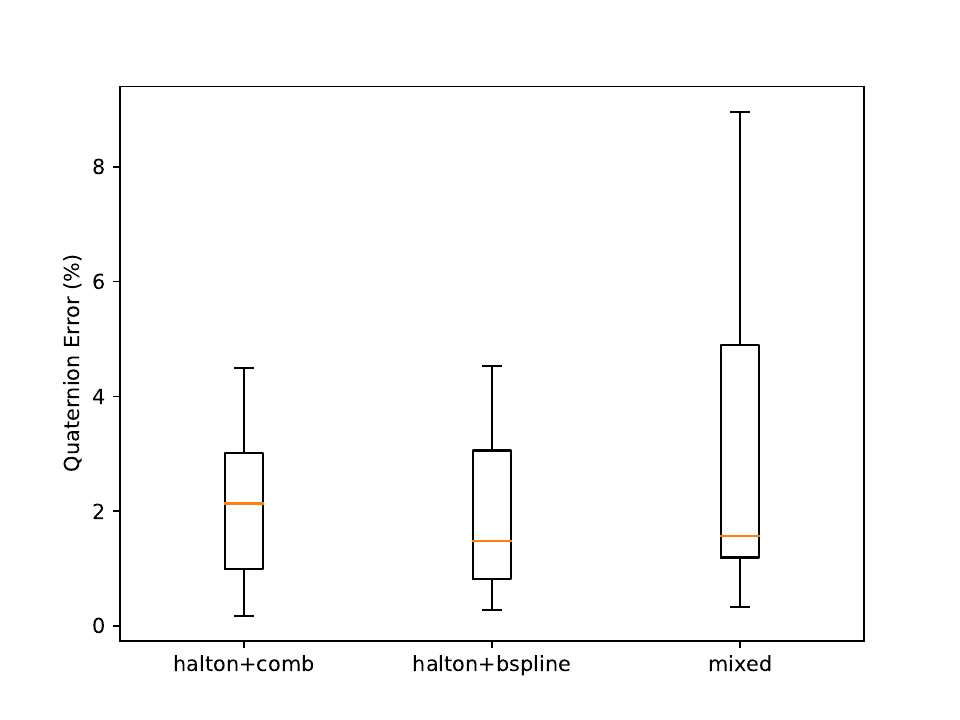}
        \caption{Quaternion Error}
        \label{fig:comb_bspline_mixed_quat_error}
    \end{subfigure}
    \begin{subfigure}[b]{0.32\textwidth}
        \centering
        \includegraphics[width=\textwidth]{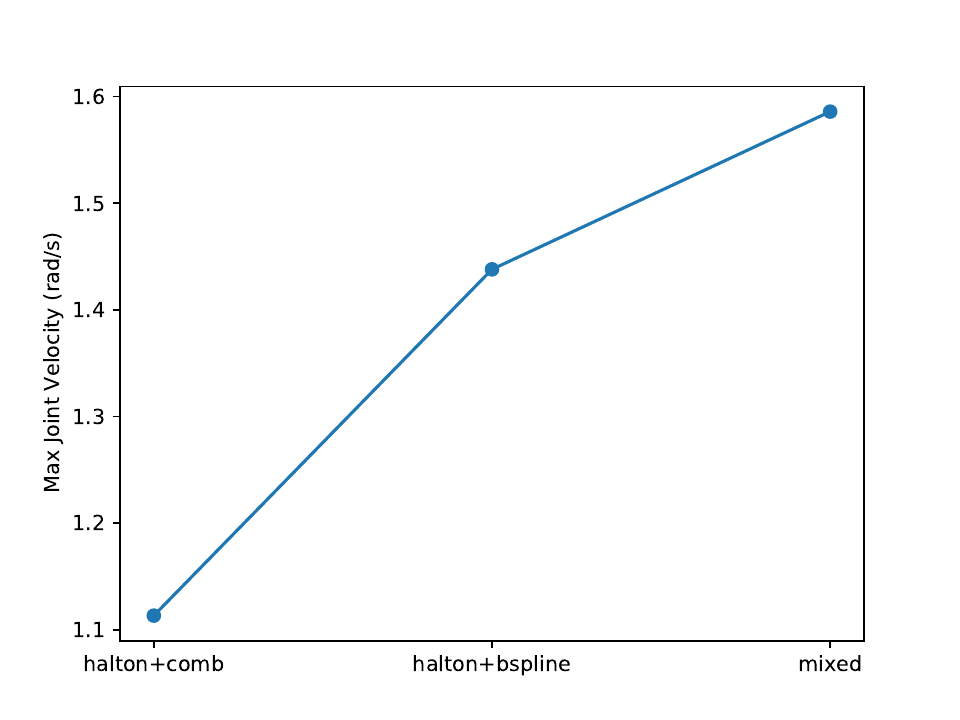}
        \caption{Max Joint Velocity}
        \label{fig:comb_bspline_mixed_max_vel}
    \end{subfigure}
    \caption{Results for the comparison of pose reaching accuracy and max joint velocity achieved for different smoothing techniques in conjunction with Halton sampling.}
    \label{fig:haltoncomb_vs_bspline}
\end{figure}

\begin{figure}
    \centering
    \begin{subfigure}[b]{0.32\textwidth}
        \centering
\includegraphics[width=\textwidth]{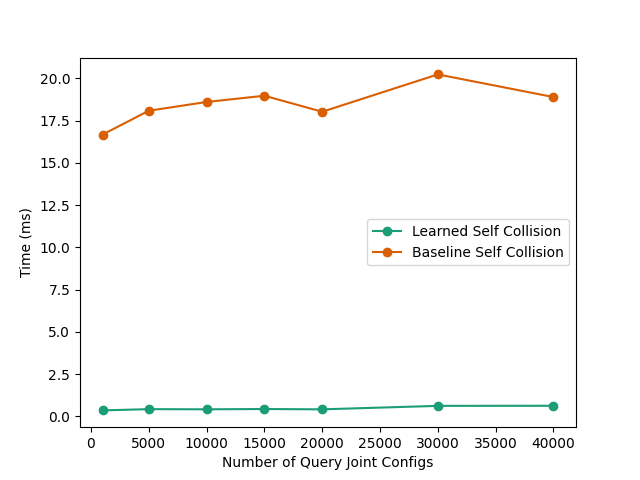}
        \caption{Timing benchmark of learned versus baseline self-collision.}
        \label{fig:learned_versus_baseline_self_collision}
    \end{subfigure}
    \hfill
    \begin{subfigure}[b]{0.32\textwidth}
        \centering
        \includegraphics[width=\textwidth]{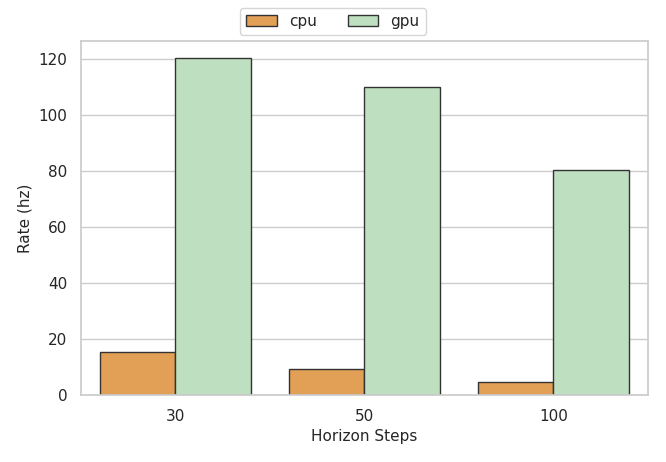}
        \caption{Timing benchmark of tensorized GPU model against CPU baseline as a function of horizon with 500 particles.}
        \label{fig:benchmark_horizon}
    \end{subfigure}
    \begin{subfigure}[b]{0.32\textwidth}
        \centering
        \includegraphics[width=\textwidth]{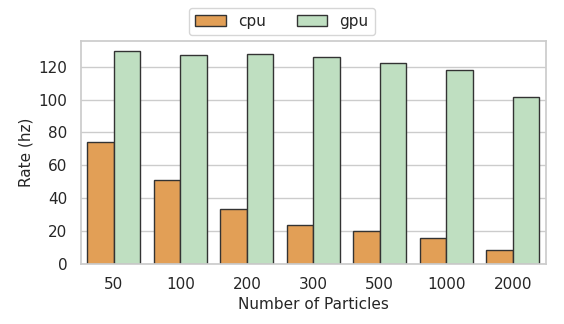}
        \caption{Timing benchmark of tensorized GPU model against CPU baseline as a function of number of particles with horizon of 30 steps.}
        \label{fig:benchmark_particles}
    \end{subfigure}
    \caption{Timing benchmarks of different key components}
    \label{fig:haltoncomb_vs_bspline}
\end{figure}

\subsubsection{Comb Filtering v/s B-Splines}
Fig.~\ref{fig:haltoncomb_vs_bspline} shows a comparison of using comb filtering versus bsplines in conjunction with Halton sampling. Comb filtering smoothens out the sampled trajectories using user defined filtering coefficients. In order to ensure smooth (low jerk) motions a very strong filtering is required which prevents the robot from ramping up this velocity. Fitting B-Splines to sampled actions is able to ensure smooth acceleration profiles which allows the robot to smoothly ramp up the velocity but comes at the price of reduced pose accuracy. However, since our sampling strategy allows us to arbitrarily mix different kinds of trajectories, we create and test a hybrid sampling strategy (denoted as "mixed") with a ratio of 0.6 and 0.4 of B-Spline and comb filtering. This provides comparable accuracy to comb filtering while also achieving high joint velocities.

\subsection{Timing Benchmarks}
\subsubsection{Learned v/s Baseline Self Collision Detection}
We quantify the computational gains obtained by using a learned self collision detector.  In Fig.~\ref{fig:learned_versus_baseline_self_collision} we present a timing benchmark for an increasing batch size of query configurations. The learned function is over 40x faster on average than baseline self collision detection that uses forward kinematics to compute link poses and calculates minimum distance between them. Further, the learned self-collision detector maintains a very low latency of 0.4-0.6ms even for large batch sizes.

\subsubsection{Speedup from Tensorized Forward Model on GPU}
The timing benchmark in Fig.~\ref{fig:benchmark_horizon} and Fig.~\ref{fig:benchmark_particles} show the computational gains from our tensorized GPU implementation of forward model versus a CPU baseline for varying horizon and number of particles used in MPC respectively. Fig.~\ref{fig:benchmark_horizon} shows that our model is over 5x faster than CPU baseline for different horizons with 500 particles. In Fig.~\ref{fig:benchmark_particles}, we can see that the our model maintains similar latency for increasing number of particles owing to it being completely tensorized.



\end{document}